\documentclass[11pt]{article}
\usepackage[utf8]{inputenc}
\usepackage[a4paper, total={6.5in, 9in}]{geometry}
\usepackage{geometry} 

\usepackage{graphicx}
\usepackage{mathptm}
\usepackage{pstricks}
\usepackage{pst-node}
\usepackage{amsmath}
\usepackage{authblk}
\usepackage{url}
\usepackage{tikz}
\usepackage{booktabs}
\usepackage[numbers]{natbib}
\usepackage{multirow}
\usepackage{caption}
\usepackage{caption} 
\usepackage{lipsum}
\usepackage{rotating}
\usepackage{wrapfig}
\usepackage[ruled,vlined,linesnumbered]{algorithm2e}
\usepackage{changepage}
\usepackage{subfig}

\providecommand{\keywords}[1]
{
  \textbf{\textit{Keywords---}} #1
}

\usetikzlibrary{decorations.pathreplacing}

\usepackage{eurosym}

\usepackage{color}

\newcommand{\mipeins}           {\texttt{m1}}
\newcommand{\mipzwei}          {\texttt{m2}}
\newcommand{\mipdreiperson}       {\texttt{m3}}
\newcommand{\mipvier}       {\texttt{m4}}
\newcommand{\mipvierb}       {\texttt{m4b}}
\newcommand{\mipvierVI}       {\texttt{m4-VI}}
\newcommand{\mipvierVIBnCbio}       {\texttt{m4bVIBnCBiO}}
\newcommand{\mipvierVIBnC}       {\texttt{m4bVIBnC}}

\newcommand{\costfirst}       {\texttt{cost}}
\newcommand{\preffirst}       {\texttt{pref}}

\title{\vspace{2cm} \Large \textbf{The bi-objective multimodal car-sharing problem}}
\bigskip
\author[a,b]{\bf Miriam Enzi}
\author[a]{\bf Sophie N. Parragh}
\author[c,d]{\bf Jakob Puchinger}
\bigskip
\affil[a]{Johannes Kepler University Linz, Institute of Production and Logistics Management, Linz, Austria \protect\\
\texttt{\{miriam.enzi,sophie.parragh\}@jku.at}
\medskip }
\affil[b]{AIT Austrian Institute of Technology, Dynamic Transportation Systems, Center for Mobility Systems, Vienna, Austria \protect
\medskip }
\affil[c]{Universit\'{e} Paris-Saclay, CentraleSup\'{e}lec, Laboratoire G\'{e}nie Industriel, Gif-sur-Yvette, France \protect
\medskip }
\affil[d]{Institut de Recherche Technologique SystemX, Palaiseau, France \protect\\
\texttt{jakob.puchinger@irt-systemx.fr}
\medskip }

\begin{document}

\begin{titlepage}
\maketitle
\thispagestyle{empty}
\end{titlepage}

\section*{Abstract}
The aim of the bi-objective multimodal car-sharing problem (BiO-MMCP) is to determine the optimal mode of transport assignment for trips and to schedule the routes of available cars and users whilst minimizing cost and maximizing user satisfaction. 
We investigate the BiO-MMCP from a user-centred point of view. As user satisfaction is a crucial aspect in shared mobility systems, we consider user preferences in a second objective. Users may choose and rank their preferred modes of transport for different times of the day. In this way we account for, e.g., different traffic conditions throughout the planning horizon.

We study different variants of the problem. In the base problem, the sequence of tasks a user has to fulfill is fixed in advance and travel times as well as preferences are constant over the planning horizon. In variant 2, time-dependent travel times and preferences are introduced. In variant 3, we examine the challenges when allowing additional routing decisions. Variant 4 integrates variants 2 and 3. For this last variant, we develop a branch-and-cut algorithm which is embedded in two bi-objective frameworks, namely the $\epsilon$-constraint method and a weighting binary search method. Computational experiments show that the branch-and cut algorithm outperforms the MIP formulation and we discuss changing solutions along the Pareto frontier.

\bigskip
\keywords{car-sharing, mobility, transportation, bi-objective, branch-and-cut}

\section{Introduction}
\label{intro}

Today, most of the world's population lives in urban environments and cities continue to grow \cite{worldpop}. Urban mobility is therefore a key topic for a sustainable future. When considering a city's infrastructure, the available mobility offers are plentiful. Public transportation provides efficient connections, some commuters use their car, others prefer bikes, scooters or even taxi. 
Besides, a trend towards sharing is clearly visible in mobility (DriveNow, Uber, ...). In short, mobility as we use it and see it is changing. 
This comes with a whole new stream of optimization problems. Only recently \citet{Mourad2019} provided a survey on the vast topic of optimizing shared mobility. 

The (privately owned) car is diminishing as the prevailing mode of transport in urban areas \cite{vcofact}.
In Vienna (Austria), the number of cars per capita is constantly decreasing \cite{pkw-bestand19, population-wien}.
People prefer other modes of transport (MOT). The modal split of cars shrank from 31\% to 25\% within the last decade. Within the same time period, bikes, public transportation and walking increased their modal split by 2 percentage points to 7\%, 38\%, and 30\%, respectively \cite{modalsplit10,modalsplit19}.
Thus, people move to alternative, more environmentally friendly MOTs. 

Additionally, citizens increasingly use sharing systems \cite{vcofact}. In Germany, the number of shared cars has increased fivefold within ten years, there are almost twelve times more users than a decade ago \cite{carsharede}.
In Vienna (Austria), one shared car eliminates the need of five privately owned ones \cite{carsharingstudie}.
At maximum 10\% of the cars in Austrian households simultaneously drive on the roads. Many car owners use their vehicles only a couple of times per year. In Lisbon (Portugal), only 3\% of the cars will be needed if all trips are covered by car- and ride-sharing. 95\% parking space can be freed up \cite{vcofact}. Moreover, car-sharing saves up to 44 million car-kilometers in Vienna annually. This equals to approximately 7,000 tons of CO$_2$ \cite{carsharingstudie}. 
Hence, by using car-sharing, resources can be employed more efficiently, it is more environmentally friendly, and newly available space can be gained as, e.g., green space in urban areas \cite{vcofact}.

The importance of rethinking mobility is clearly visible in the presence of prominent concepts in various cities. 
Vienna targets a split of 80:20 where 20\% of the trips are covered by cars, the others by public transportation, bikes or walking. The idea is to extend the mobility offers with profound sharing concepts and to move towards the vision of a 'Smart City' \cite{step2025}.
Madrid is aiming to establish a holistic 'Mobility as a Service' concept offering real-time information and including over 30 shared mobility options \cite{madrid}.
Within novel mobility concepts, bikes are receiving exceptional attention. Vienna almost doubled the cycling network in the last decade, and accomplished a similar increase in kilometers driven on specific legs \cite{fahrradwien, entwicklungFahrrad}.
Paris presents the 'Plan V\'{e}lo' where the target is to emerge to the world's bike capital. The ambition is to minimize the space for cars and make space for bike usage and pedestrians \cite{planvelo}.

Novel mobility concepts and reconsidering mobility plays an important role not only in a private environment, but also in a corporate setting. Companies increasingly aim to provide mobility concepts for their employees. {This work is part of an applied research project SEAMLESS (http://www.seamless-project.at), in which project partners, such as the \textit{Austrian Post AG} or \textit{T-Systems Austria GesmbH}, strive for the implementation of the discussed ideas.}  The target is to reduce {a} one-to-one assignment of company cars, employ more environmentally friendly MOTs and strive for shared mobility where each employee gets her preference. This goes hand in hand with companies aiming for a greener carbon footprint and enhancing employee satisfaction \cite{seamless}. 

Traveler experience needs to be taken into account in the design of novel mobility systems and is key to its success with users \cite{Ouail2019}. Thus, when studying mobility, convenience and user preferences are crucial. 
However, from an operator perspective the cost-factor plays an important role as well and usually cost-efficiency is in conflict with a MOT's convenience. This 'convenience' is difficult to measure and must be tackled on an individual user level. As we observe, and also other authors studying mobility have outlined \cite{Ferrero2018}, including user preferences can decide on the 'win or lose' of a system. Therefore, we investigate the trade-off between minimizing cost and enhancing the individual satisfaction of a user in a mobility system. Combining these parameters and providing the decision maker with a set of efficient solutions will lead to an enhanced acceptance of such a system.

Motivated by this, we study the bi-objective multimodal car-sharing problem where we assign MOTs to trips and find car and, depending on the variant, also user routes throughout a day. 
We formulate two objectives to minimize cost and maximize user satisfaction. We further take into account the possibility of variation of user preferences and travel times throughout the day, becoming time dependent input parameters. 
{We refer to car-sharing throughout the paper as to where a group of users is mutually using a pool of cars.}
{Note that the output aims to provide an optimal assignment of MOTs throughout a day using time-dependent travel times.}

Our main contributions are:
\begin{itemize}
	\item 
	We introduce the bi-objective multimodal car-sharing problem (BiO-MMCP). We present four variants of the problem, discussing increased flexibility of the timings of the visits: we present the model (i) with fixed task sequences and without time-dependent travel times and user preferences, (ii) with fixed time sequences and including time-dependent {travel} times and user preferences, (iii) no fixed sequences and no time-dependent {travel} times or preferences, and lastly (iv) open sequences of tasks and time-dependent travel times and user preferences. 
	\item
	We propose a branch-and-cut algorithm for the most complex problem variant examined in this paper. The algorithm is embedded into two bi-objective frameworks, namely the $\epsilon$-constraint method and a weighting binary search method. We show that for both frameworks it is highly beneficial to add cuts in the form of constraints from prior iterations to the following iterations.
	\item
	We provide a thorough analysis where we (i) compare different solution approaches for the models, and (ii) give insights into the trade-offs between cost-minimization and enhancing user-centred MOT preferences.
\end{itemize}

The paper is organized as follows:
In Section~\ref{sec:relatedwork} we review related work. 
Section~\ref{sec:bio-mmcp} introduces the BiO-MMCP where
Section~\ref{sec:problemformulation} gives a problem formulation, followed by the formal description in Section~\ref{sec:formal} for all four variants. 
In Section~\ref{sec:solutionmodel} we describe the implemented solution approach. As most of the variants are solved as a mixed integer program (MIP) with the generic MIP solver CPLEX, we focus on the branch-and-cut developed for the last variant of the model, described in Section~\ref{sec:branchncut}. Moreover, we introduce a set of valid inequalities in Section~\ref{sec:validineq} and describe the bi-objective frameworks used in this paper in Section~\ref{sec:bioframework}.
Section~\ref{sec:comp} summarizes our computational study.
Finally, we draw conclusions and we give an outlook to future work in Section~\ref{sec:concl}.

\section{Related work}
\label{sec:relatedwork}

Research addressing the design and implementation of car-sharing systems has risen over the past years. Many existing papers focus on strategic decision making, such as the design of services, infrastructure (e.g. design/location of facilities or charging stations) or fleet management. Nevertheless, various papers stress the importance of integrating the user related attributes in optimization problems tackling sharing systems. A comprehensive literature review has been presented by \citet{Ferrero2018}.

A large amount of research has been performed on data collection, data analysis and simulation based studies in order to assess the potential impacts of car-sharing systems. Most of these studies have been conducted on city-wide public systems. Demand for car-sharing systems and impacts on mobility behaviour are typically assessed through questionnaires~\cite{Zhou2011, Sioui2013}. The potential and effects of such systems are then often determined through simulation based approaches~\cite{Ciari2014}. From an operational perspective problems considered in the car-sharing related literature are mainly concerned with relocating, recharging and servicing vehicles~\cite{Almeida2012, Nair2014, Weikl2013}.
The problem we are introducing in this article however, is an operational problem for planning trips and allocating means of transport in a closed system where travel demand is known in advance. Embedding car-sharing in a multimodal system and especially treating it in a bi-objective formulation is a novel way of addressing car-sharing from a user-centered perspective.

In a different line of research, ride-sharing has attracted an increased amount of interest in the last years. Major research efforts have been made in analyzing and designing such services. Strategic and tactical decisions as well as the development of new algorithms for daily operations have also been in focus of recent work. A comprehensive survey on such approaches can be found in~\citet{Mourad2019}. A large number of case studies mainly based on simulation and data analysis have been published on the potential impact and feasibility of various sharing schemes with a focus on ride-sharing~\cite{Wolfler2004, Ferreira2014, Maciejewski2016, Tachet2017}.

For the first two variants of our proposed problem, where the task sequence is fixed, we refer to the Fixed Sequence Arc Selection Problem (FSASP) which was introduced by \citet{Garaix2010} and proven to be NP-hard. The FSASP considers a fixed sequence of nodes that are linked by multiple arcs. Choosing an arc between two nodes is the subject of determination. This problem applies to the first two variants of our problem in this paper. Note that only recently \citet{Huang2017} shortly stressed that this research direction can be a good basis for further algorithmic work, naming home appliance delivery companies as an example. 
As we additionally determine the sequence of visited nodes, we can detect similarities to the VRP \cite{Toth2002, Eksioglu2009} in our work. Our paper introduces a kind of multi-trip VRP \cite{Cattaruzza2016} with heterogeneous vehicles and multiple depots on a multi graph. 
\citet{Garaix2010} were among the first who studied VRPs with alternative arcs between each pair of nodes. VRPs with multiple attributes \cite{Garaix2010} or multi-graphs in the VRP stream have gained increasing attention in the past years \cite{Doppstadt2016, BenTicha2017, BenTicha2018, BenTicha2019, Huang2017}, whereas, of course, mutlimodality significantly enlarges the set of possible solutions \cite{Caramia2009}. Research considering various attributes on arcs is fairly recent, yet highly important to consider as one connection of nodes usually implies specific trade-offs (usually time vs. cost) which are not considered on a classical graph. We consider the characteristics of different modes of transport as well as time-dependent preferences and costs jointly on one arc.
We refer to \citet{Gendreau2015}, for a review on time-dependent routing problems. However, we could not find any work introducing time-dependent preferences on modes of transport in a car-sharing context.

Integrating customer-oriented aspects into optimization problems, or more specific vehicle routing problems, is a topic of increasing interest. In \citet{Vidal2019} a detailed analysis through VRP variants also tackling customer-centred objectives is provided. As an example, the cumulative VRP \cite{Ngueveu2010, Silve2012} replaces the classical minimum cost objective function with the minimization of individual customer arrival times. 
\citet{Martinez-Salazar2015} introduce a customer-centric multi-trip VRP with a single vehicle minimizing the sum of customer waiting times to receive a specific service. On a somewhat different but related topic, \citet{Braekers2016} introduce a bi-objective routing and scheduling problem for home care where the second objective minimizes client inconvenience. In our work, we optimize user preferences for MOTs as a second objective function.
\citet{Jozefowiez2008} review numerous papers tackling multiple objectives in the context of VRPs. They name the most common objectives to be cost, length of the tour, balance or problem specific objectives. Since then, various papers have been published. Recently it seems that there is a vast amount of published research with environmental \cite{Kargari2018, Alexiou2015, Konstantinos2017, Demir2014, Eskandarpour2019, Ghannadpour2019, Toro2017, Tricoire2017,Govindan2019,Anderluh2019,Grabenschweiger2018} or external social criteria \cite{Ghannadpour2019,Govindan2019,Nolz2014,Anderluh2019,Grabenschweiger2018} as alternative objectives.

Multi-objective optimization gives a deeper insight into the solution pool of a problem. However, there might exist a large number of trade-off solutions. The target is to find an efficient set of solutions that cannot be optimized in one objective without worsening another one. Those efficient solutions are then called Pareto optimal solutions. There is a vast amount of works on exact as well as heuristic approaches to solve for multicriteria optimization problems. 
Prevailing metaheuristics in this field are evolutionary algorithms such as the NSGA-II \cite{Deb2000} or the SPEA-II \cite{Gharari2016}. However, only recently \citet{Matl2019} have shown that single-objective VRP heuristics can be efficiently used in an $\epsilon$-constrained-based method. The $\epsilon$-constraint method \cite{Haimes1971,Srinivasan1976} is one of the prevailing methods to solve multi-objective optimization problems. It repeatedly solves a single-objective optimization problem by considering the other objectives in terms of constraints.
Further widely applied frameworks to solve multi-objective problems are the two-phase method \cite{Visee1998}, the weighted sum approach \cite{Aneja1979} or, more recently, the balanced box method \cite{Boland2015} and the weighting binary search method \cite{Riera2005}.  These so called criterion space methods, embed a single-objective optimization problem and systematically enumerate the Pareto frontier. 
However, recent works focus on adapting the branch-and-bound algorithm to solve the multi-objective case in a single run \cite{Stidsen2014,Vincent2013,Parragh2019,Adelgren2017}. 
A recent overview of exact methods for multi-objective optimization is provided in \citet{Ehrgott2017}. A detailed overview of general multi-objective combinatorial optimization is provided by \citet{Ehrgott2003}. 
For our study we choose the $\epsilon$-constraint method as well as a weighting binary search as they are relatively simple to implement and have shown to be very efficient. The latter one is based on the algorithm proposed by \citet{Riera2005}, who developed a weighting method and conduct a binary search in the objective space. 
Moreover, similar to \citet{Berube2009}, we use a branch-and-cut approach relying on previous information for subsequent problems by adding cuts to the subproblem. 
Similarly in \citet{Riera2005} cuts from prior iterations are added to the cut pool for further iterations. 
Contrary to \citet{Riera2005} and \citet{Berube2009}, we add detected cuts as hard constraints, showing better results for our problem setting.

\section{The bi-objective multimodal car-sharing problem}
\label{sec:bio-mmcp}

In the following we describe the BiO-MMCP and give a formal description of the variants of the problem studied in this paper.

\subsection{Problem description}
\label{sec:problemformulation}

The BiO-MMCP aims to assign modes of transport to user trips and determining car routes during a day while minimizing cost and maximizing user satisfaction by accounting for MOT preferences. 

Each user trip starts in a depot, covers a set of tasks and ends in a depot again. A user may have more than one trip during a day. A route is a sequence of trips during a day. Note that we introduce car routes and user routes: A car route schedules the trips covered by one car during a day, whereas the car is handed over at the depot from one user to another. A user route consists of all the trips assigned to one user during a day, whereas the user may change MOTs between trips (i.e. at the depot). 

We consider a closed group of users and a set of possible MOTs. A pool of cars is given and all other MOTs are considered to have infinite capacity. {With this assumption we are able to cover all demanded trips. This also has practical implications as, e.g., there is usually no spatial or temporal limit on the availability of public transport in a city during a day. This also holds for bikes, as due to several bike-sharing offers, we can assume that bikes are available at any time in a city.} Each user may give preference scores to the available MOTs where we assume the lower the score the better the MOT is rated (scale 1-10 where 1 is best). Moreover, depending on the variant of the problem, users may determine preferences for different times of the day, resulting in time-dependent user-based MOT preferences. Furthermore, we introduce time-dependent travel times as, e.g., the car drive will take longer through rush-hour than at noon. As our cost function also comprises cost of time, the adapted travel times will have an impact on the cost function. {Note that even though travel times may be stochastic, we can plan within a deterministic setting as we use time-dependent travel times for all modes of transport.}

The goal of the BiO-MMCP is to cover a set of trips for a given planning horizon by assigning MOTs to trips and determine car routes (optionally also user routes) for a closed community. The locations of the start and end points as well as the tasks of a trip are fixed. This means, it is known in advance which user will visit which task. Depending on the considered variant of the problem, the sequence of the tasks may vary.

We investigate four variants of the introduced problem: 

\paragraph{Model 1 (\mipeins)} In the first variant we assume that each user follows a fixed sequence of tasks, starting and ending at a fixed (but possibly different) depot. Preferences are given for each MOT for each user. We aim to find the best MOT to trip assignment and to determine the car routes. The objectives are to minimize costs and MOT preferences. In this variant, user routes are assumed to be given.

\paragraph{Model 2 (\mipzwei)} In this variant we assume the same setting as in model \mipeins\ but include time-dependent MOT preferences and travel times. The target is to find the best MOT to trip assignment and schedule the car routes from a pool of cars whilst minimizing time-dependent costs and user preferences. Again, user routes are input to the problem.

\paragraph{Model 3 (\mipdreiperson)} In the third variant we consider a fixed user to tasks assignment, and start and end locations. However, the sequence of tasks within a trip as well as the sequence of user trips are subject of determination. This means that we have to, in addition to car routes, find user routes throughout a day. The objectives are again to minimize costs and user preferences.

\paragraph{Model 4 (\mipvier)} This model is a combination of model \mipzwei\ and \mipdreiperson: we consider time-dependent user preferences and travel times as well as variable task and trip sequences. Thus, we intend to determine the MOT assignment, schedule car as well as user routes whilst minimizing both time-dependent MOT preferences of users and costs.


\subsection{Formal description}\label{sec:formal}

We now formally introduce the different variants and their respective mathematical formulations, using the following notation (also summarized in Table~\ref{tab:notation}):

\begin{table}
\caption{Mathematical notation used in the formal description of the BiO-MMCP.}
\label{tab:notation}
\hrule{\smallskip}
    \small
    \begin{tabular}{r|l}
    \multicolumn{2}{l}{\textbf{ Sets and nodes}} \\
    $P$ & set of users \\
    $R$     &   set of trips \\
    $Q_r \subseteq Q$ & set of tasks of trip $r$ \\ & as a subset of the set of tasks \\
    {$a_r$} & {start node of a trip $r$} \\
    {$b_r$} & {end node of a trip $r$} \\
    $G_r$    &   set of nodes on a trip $r$ \\
    $D$ & set of depots \\
    $K$ & set of modes of transport \\
    {$L$} & {set of legs} \\ 
    $L^-,L^+$  & set of ingoing/outgoing legs \\
    {$L_v^-,L_v^+$ }  & {set of ingoing/outgoing legs of node $v$} \\
    {$L_d^-,L_d^+$ }  & {set of ingoing/outgoing legs of depot $d$} \\
    {$L_{p}$}  & {set of legs assigned to user $p$} \\
    {$L_r$} & {set of legs on a trip $r$} \\
    {$L_{vp}$ }& {set of legs of a user $p$ going} \\ & {in/out of a node $v$} \\
    {$L_{vk}^-,L_{vk}^+$ }  & {set of ingoing/outgoing legs} \\ & {of node $v$ by MOT $k$} \\
    {$L_{vp}^-,L_{vp}^+$ }  & {set of ingoing/outgoing legs} \\& {of node $v$ by user $p$} \\
    {T} & {set of time periods} \\
    {$S$} & {subset of the set of nodes $G_r$ of a trip $r$ }\\
    $A_p \subseteq A$  & set of trip start nodes of a user $p$ \\ & as a subset of the set of trip \\ & start nodes \\
    $B_p \subseteq B$  & set of trip end nodes of a user $p$ \\ & as a subset of the set of trip \\ & end nodes \\
    $V$    &  set of all nodes \\
    $V'$    &  set of nodes without depots, V $\setminus$ D \\
    $T$ &   set of time periods \\
    $\mathcal{R}$ &   set of infeasible paths \\
    {$\mathcal{R}_{car}$} & {set of infeasible car routes} \\
    {$\mathcal{R}_{p}$} & {set of infeasible user routes} \\
    $\gamma_p$ & start node of person $p$ \\
    $\phi_p$ & end node of person $p$ \\
\end{tabular}
    \begin{tabular}{r|l}
    \multicolumn{2}{l}{\textbf{Input parameter}} \\
    $W_{dk},\overline{W}_{dk}$     & number of MOTs $k$ in depot $d$ at \\ & beginning/end of \\ & the planning horizon \\
   { $\sigma_{pk},\sigma^t_{pk}$ }& preference value of a person $p$ \\ & for MOT $k$ (for time period $t$) \\
      $\theta_{l}$    & preference of leg $l$ \\
      $y_l$ & origin of leg $l$ \\
      $z_l$ & end of leg $l$ \\
      $c_{l}$   & cost of leg $l$ \\
      $u_l$ &   user of leg $l$ \\
      $m_l$ &   MOT of leg $l$ \\
      $h$   &   maximal waiting time \\
      $H$ & end of planning horizon \\
      $M$ & big M \\
      $t_l$   &  driving time of leg $l$ \\
      $s_v$   & service time at node $v$ \\
      $o_l$   & interval start of leg $l$ \\
      $e_l$   &  interval end of leg $l$ \\
      {$\mathcal{W}$} & {accumulated waiting time} \\
      {$\Delta$} & {value stating how much a route} \\ & {can be moved forward} \\
      {F} & {forward slack, $F=\mathcal{W}+\Delta$} \\
     \noalign{\smallskip}\hline\noalign{\smallskip}
      \multicolumn{2}{l}{\textbf{Decision variables}} \\
      $x_{l}$ & 1 if leg $l$ is chosen, 0 otherwise \\
       $\tau_l$    & departure time of leg $l$ \\
       \multicolumn{2}{l}{} \\
       \multicolumn{2}{l}{} \\
       \multicolumn{2}{l}{} \\
       \multicolumn{2}{l}{} \\
       \multicolumn{2}{l}{} \\
       \multicolumn{2}{l}{} \\
       \multicolumn{2}{l}{} \\
       \multicolumn{2}{l}{} \\
       \multicolumn{2}{l}{} \\
       \multicolumn{2}{l}{} \\
       \multicolumn{2}{l}{} \\
    \end{tabular}
\hrule{\smallskip}
\end{table}

Given is a set of users $P$ and a set of trips $R$, where each trip $r \in R$ has a set of tasks $Q_r$ assigned. A trip starts in a depot $a_r$, ends in depot $b_r$ and covers in between one or more tasks $q$. We store all nodes assigned to a trip $r$ in the set $G_r$, where $r = \{a_r,q^r_1,q^r_2,...,b_r\}$. Note that a user $p$ might cover more than one trip during a day. The set of tasks $Q_r$ is known in advance whereas each task $q$ is unique and may only be in one set $Q_r \subseteq Q$, where $Q$ denotes the set of all tasks. We model the connections between two subsequent tasks as a leg $l$.

Furthermore, we consider a set of depots $D$, which are artificial nodes representing start/end points of car routes during a day, i.e. each route starts and ends here. The start depot $d$ is connected to all starting nodes $a$, and conversely each end node $b$ is connected to the end depot $d'$. 

We consider a set of modes of transport $K = \{car,public,bike\}$, where $public$ comprises public transportation including walking. If a trip starts by a MOT, then the MOT will be used for the full trip. We assume at each depot $d \in D$ an available number of MOTs $k$ at the beginning and end of the planning horizon, denoted as $W_{dk}$ and $\overline{W}_{d'k}$, respectively. 

We denote the set of all nodes by $V$ and $V'$ be the set of nodes without the set $D$, such that $V' = V \setminus D$. For every node $v \in V$ we have the set of outgoing legs $L_{vk}^+$ and ingoing legs $L_{vk}^-$ by MOT $k$. All legs are stored in the set of all legs $L$. We store any relevant information on the legs.

Each user $p$ assigns a preference value $\sigma_{pk}$ to each of the given modes of transport $k \in K$. Note that, as we also minimize the preference objective, we assume that the lower the score, the better the user values the mode of transport. As a leg $l$ refers to exactly one mode of transport $k$ and one user $p$, we assign the value $\sigma_{pk}$ to the respective leg $l$, denoted as $\theta_l$.
The cost value $c_l$ of a leg $l$ consists of variable distance cost, cost of time and cost of emissions. For more information, we refer to Section~\ref{sec:data}. 

For time-dependent user preferences we define a set of time periods $t \in T$ during the day. A time period replicates, e.g., rush-hours. Each user $p$ determines a preference value $\sigma^t_{pk}$ for each of the given time periods $t$ and MOT $k$. In the case when a leg $l$ completely lies within a period $t$ the preference value of the leg $\theta_{l}$ equals $\sigma^t_{pk}$. In the case where the leg covers more than one period, we calculate a weighted average of the preference values. 
As our cost also depends on time, we also adapt the cost term considering time-dependencies in the same way. 

\begin{figure}
\subfloat[Trip $r$ with its associated legs $l$, and the respective cost and preference values given as ($c_l,\theta_l$).]{
\begin{tikzpicture}[scale=0.75]
\path (0,5)  node[draw,shape=rectangle] (depot3) {$a$};
\path (15,5) node[draw,shape=rectangle] (depot5) {$b$};
\path (5,5) node[draw,shape=circle] (10) {$q_0$};
\path (10,5) node[draw,shape=circle] (4) {$q_1$};
\draw[->]  (depot3) to node [above, sloped] (TextNode1) {(10,2)} (10);
\draw[dotted, ->]  (depot3) to [bend right=30] node [below, sloped] (TextNode1) {(12,3)} (10);
\draw[dashed, ->]  (depot3) to [bend left=30] node [above, sloped] (TextNode1) {(15,4,)} (10);
\draw[->] (10) to node [above, sloped] (TextNode1) {(20,2)} (4);
\draw[dotted, ->] (10) to [bend right=30] node [below, sloped] (TextNode1) {(40,3)} (4);
\draw[dashed,->] (10) to [bend left=30] node [above, sloped] (TextNode1) {(30,4)} (4);
\draw[->] (4) to node [above, sloped] (TextNode1) {(10,2)} (depot5);
\draw[dotted, ->] (4) to [bend right=30] node [below, sloped] (TextNode1) {(12,3)} (depot5);
\draw[dashed,->] (4) to [bend left=30] node [above, sloped] (TextNode1) {(15,4)} (depot5);
\end{tikzpicture}
}
\\
\subfloat[Trip $r$ with its associated legs $l$, and the respective time-dependent cost and user-preferences for each time period $t$ given as $(\lbrack c^{t_0}_l,c^{t_1}_l,c^{t_2}_l\rbrack),\lbrack \theta^{t_0}_l,\theta^{t_1}_l,\theta^{t_2}_l \rbrack$.]{
\begin{tikzpicture}[scale=0.75]
\path (0,5)  node[draw,shape=rectangle] (depot3) {$a$};
\path (15,5) node[draw,shape=rectangle] (depot5) {$b$};
\path (5,5) node[draw,shape=circle] (10) {$q_0$};
\path (10,5) node[draw,shape=circle] (4) {$q_1$};
\path (0,8)  node (16) {};
\path (2,7.5)  node (xx) {$t_0$};
\path (0,2)  node (17) {};
\path (4,8)  node (20) {};
\path (7.5,7.5)  node (xxx) {$t_1$};
\path (4,2)  node (21) {};
\path (11,8)  node (24) {};
\path (13,7.5)  node (x) {$t_2$};
\path (11,2)  node (25) {};
\path (15,8)  node (28) {};
\path (15,2)  node (29) {};
\draw[->]  (depot3) to node [above, sloped] (TextNode1) {([{\color{red} 14},{\color{red} 12},10],[{\color{red} 2},{\color{red} 1},1])} (10);
\draw[dotted, ->]  (depot3) to [bend right=30] node [below, sloped] (TextNode1) {([{\color{red} 8},{\color{red} 10},8],[{\color{red} 3},{\color{red} 4},4])} (10);
\draw[dashed, ->]  (depot3) to [bend left=30] node [above, sloped] (TextNode1) {([{\color{red} 16},{\color{red} 13},15],[{\color{red} 4},{\color{red} 3},5])} (10);
\draw[->] (10) to node [above, sloped] (TextNode1) {([23,{\color{red} 22},20],[2,{\color{red} 1},1])} (4);
\draw[dotted, ->] (10) to [bend right=30] node [below, sloped] (TextNode1) {([33,{\color{red} 35},33],[3,{\color{red} 4},4])} (4);
\draw[dashed,->] (10) to [bend left=30] node [above, sloped] (TextNode1) {([31,{\color{red} 28},30],[4,{\color{red} 3},5])} (4);
\draw[->] (4) to node [above, sloped] (TextNode1) {([14,{\color{red} 12},{\color{red} 10}],[2,{\color{red} 1},{\color{red} 1}])} (depot5);
\draw[dotted, ->] (4) to [bend right=30] node [below, sloped] (TextNode1) {([8,{\color{red} 10},{\color{red} 8}],[3,{\color{red} 4},{\color{red} 4}])} (depot5);
\draw[dashed,->] (4) to [bend left=30] node [above, sloped] (TextNode1) {([16,{\color{red} 13},{\color{red} 15}],[4,{\color{red} 3},{\color{red} 5}])} (depot5);
\draw[dotted,-] (16) to (17);
\draw[dotted,-] (20) to (21);
\draw[dotted,-] (24) to (25);
\draw[dotted,-] (28) to (29);
\end{tikzpicture}
}\\
\centering
\begin{tikzpicture}
\scriptsize
\draw (0,0) -- (1,0) node [midway, above, sloped] (TextNode) {car};
\draw [dashed] (2,0) -- (3,0) node [midway, above, sloped] (TextNode) {public transp.};
\draw [dotted] (4,0) -- (5,0) node [midway, above, sloped] (TextNode) {bike};
\end{tikzpicture}
\caption{Example of one trip with its associated legs $l$ starting in node $a$, visiting tasks $q_0$, $q_1$ and ending in node $b$. Between the nodes we insert different legs for each mode of transport, which are car, public transportation and bike in our case. A label of a leg is defined with two attributes: cost and preferences. Figure (a) shows a simple trip, where no time-dependencies are considered. Figure (b) includes time-dependent information for the respective periods $t$.} \label{fig:path}
\end{figure}
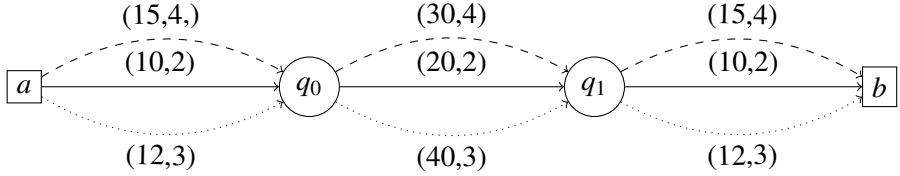
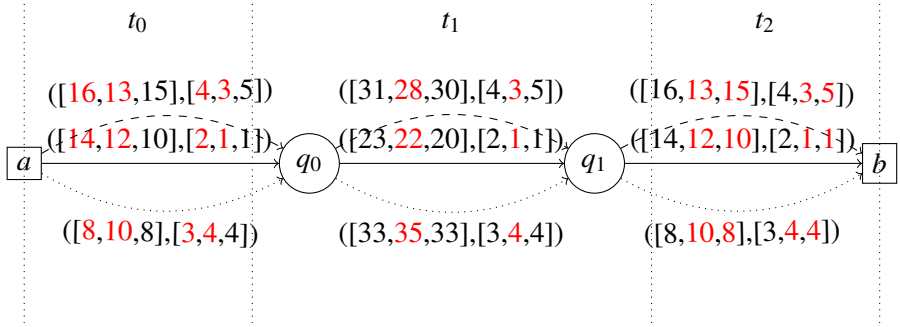

Figure~\ref{fig:path}(a) shows an example of a simple trip $r$. It starts in node $a$ and ends in $b$ whilst visiting $q_0$ and $q_1$. We insert legs for each mode of transport (denoted by different lines) between each node and assign the respective cost and preference value, given in brackets as ($c_l$,$\theta_l$). We do not consider time-dependent travel times or user preferences here.
\noindent
Figure~\ref{fig:path}(b) shows the same trip as Figure~\ref{fig:path}(a), but considers time-dependencies. Therefore, three time periods are indicated as $t_0,t_1,t_2$. For each leg we have cost and preference values for each of the respective periods. The legs between $q_0$ and $q_1$ lie completely within one time period and can therefore be taken as they are. For the others, we compute the share of each time period on the leg and get the respective preference value and cost by computing the weighted average.

\subsubsection{Model 1 (\mipeins)}

In model \mipeins\ the sequence of tasks is fixed, resulting in predetermined trips $r \in R$. 
We connect each $a$ with its fixed successor $q$, each task $q$ with its fixed successor $q'$ or, if the trip only covers one task, the trip end node $b$. For every pair of end and start nodes $(b,a)$ where $a$ is ahead in time, we insert an additional artificial leg  with costs and preferences 0, in order to allow for the connection of car routes covering more than one trip throughout the day. 

Each leg in the graph results in a tuple $\{(u_l,y_l,z_l,m_l,c_l,\theta_l)\}$ where $u_l$ is the assigned user, $y_l$ and $z_l$ are the origin and end of the leg, $m_l$ the assigned MOT, $c_l$ the cost and $\theta_l$ the preference value. 

The introduced binary decision variable $x_{l}$ takes on value 1 if leg $l$ is chosen and 0 otherwise. 

With this, we can introduce a compact formulation for the first version of the BiO-MMCP.

\begin{eqnarray}
\mbox{min}	\label{eq:mip1-OF1}
			& \quad \sum\limits_{l \in L} c_{l}  x_{l} \\
\mbox{min}	\label{eq:mip1-OF2}
			& \quad \sum\limits_{l \in L} \theta_{l}  x_{l} \\
\mbox{s.t} 
            \label{eq:mip1-coverevery}
			&\sum\limits_{k \in K} \sum\limits_{l \in L_{vk}^+} x_{l} = 1  &\forall v \in V'  \\
 			\label{eq:mip1-flowcons}
			&\sum\limits_{l \in L_{vk}^-} x_{l} = \sum\limits_{l \in L_{vk}^+} x_{l}  &\forall v \in V', k \in K  \\
			\label{eq:mip1-depotstart}
			&\sum\limits_{l \in L_{dk}^+} x_{l} \leq W_{dk}  &\forall d \in D, k \in K \\
			\label{eq:mip1-depotend}
			&\sum\limits_{l \in L_{d'k}^-} x_{l} \leq \overline{W}_{d'k}  &\forall d' \in D, k \in K  \\
			\label{eq:mip1-bins}
			& x_{l} \in \{0,1\}	&\forall l \in L
\end{eqnarray}
\noindent
The objective (\ref{eq:mip1-OF1}) minimizes total cost and objective (\ref{eq:mip1-OF2}) minimizes user-centred MOT preferences.
Constraints~(\ref{eq:mip1-coverevery}) make sure that each node $v$ is covered by exactly one leg $l$.
Constraints (\ref{eq:mip1-flowcons}) ensure flow conservation at nodes $v \in V'$ for every MOT $k$.
Constraints~(\ref{eq:mip1-depotstart}) and (\ref{eq:mip1-depotend})  restrict the number of available MOTs $W_{dk}, \overline{W}_{d'k}$ at the start and end of the time horizon.
Constraints~(\ref{eq:mip1-bins}) define the domains of the decision variables.

\subsubsection{Model 2 (\mipzwei)}

We extend the previous model by introducing time-dependent MOT preferences and costs. 
We assume fixed times of tasks $q$. With this, and as we know the driving time of a leg, we can exactly determine start and end times of the leg and thus assign a preference value.

As we store all relevant information directly on the leg $l$, we do not have to model time explicitly. 
This results in the same tuple $\{(u_l,y_l,z_l,m_l,c_l,\theta_l)\}$ as before, with a modified value of $\theta_l$ and $c_l$.
As we only have a change in the data, but the model remains unchanged, we use model \mipeins\ again.

\subsubsection{Model 3 (\mipdreiperson)}

In model \mipdreiperson\ we have to determine the sequence of tasks per user (ensuring no subtours) as well as consider the scheduling of trips each user is taking. Therefore, the underlying graph has to be adapted. 
We again consider the set of all nodes $V$, the set of intermediate nodes $V'$, the set of depots $D$, the set of MOTs $K$, the set of legs $L$, and the set of users $P$. We define sets $A_p$ and $B_p$ containing all start nodes $a$ and end nodes $b$ of a user $p$, respectively. These sets will consist of exactly one node, if a user is taking only one trip, two if the user has two trips, etc.
Previously, to assure car routes, we only connected an end node $b$ of a trip to a start node $a$ of another trip if $a$ was ahead in time of $b$. As we are not considering any fixed times/sequences anymore, we connect every $b$ to every $a$ if they are in the same physical depot. Similarly we connect all nodes belonging to one set $G_r$, yet not changing the predetermined start and end nodes of one trip. For now, we do not consider time-dependent preferences on legs.
Note that the tasks lying on a specific trip are fixed, meaning that if a user previously had two trips, the user will again cover two trips. 

In order to prevent parallel trips of one user, the user routes are modeled into the graph. Doing so, we add new artificial nodes $\gamma_p$ and $\phi_p$ for each user $p$ where the user starts and ends the respective route during a day (similar to the idea of the $d \in D$ where all MOT flows start). We connect the respective $\gamma_p$ to all start nodes $a \in A_p$ of one user $p$ and conversely the respective $\phi_p$ to all $b \in B_p$. We connect user trips by inserting a leg $l$ between $b,a$ of the same user.
Note that, instead of modifying the underlying graph, we also used additional constraints in the model. However, this formulation turned out to be very weak.

As the sequence of tasks of a trip is not fixed we determine the departure time $\tau_l$ of a leg $l$. By assuring increasing times of legs, we also avoid subtours. Additionally, in order to avoid unrealistic long waiting times at nodes, we assume that a user can wait for a maximum amount of time before she continues her trip, e.g. 30 minutes, denoted as $h$. 

Model \mipdreiperson\ can now be stated as follows, where decision variables $\tau_l$ give the departure time of leg $l$, $H$ depicts the end of the planning horizon, $M$ denotes a big M, $t_l$ is the travel time of a leg $l$ and $s_v$ the duration of the task.

\begin{eqnarray}
\mbox{min}	&\eqref{eq:mip1-OF1} \nonumber \\
\mbox{min}	&\eqref{eq:mip1-OF2} \nonumber \\
\mbox{s.t} 
            & \eqref{eq:mip1-coverevery}-\eqref{eq:mip1-bins}   \nonumber \\
            \label{eq:mip3b-subtour-elimination}
			&\tau_l + t_l + s_v - \tau_n \leq M (2 - x_l - x_n) &\forall l \in L_{vk}^-, n \in L_{vk}^+, v \in V', k \in K \\
			 \label{eq:mip3b-30min}
            & \sum\limits_{l \in L^-_v} (\tau_l + t_l x_l) + s_v \geq \sum\limits_{l \in L^+_v} \tau_{l} - h & \forall v \in V' \\
			\label{eq:mip3b-planning-horizon-end}
			&\tau_l  \leq H x_l &\forall l \in L \\
			\label{eq:mip3b-personout}
			&\sum\limits_{l \in L_{\gamma_p}^+} x_{l} = 1  & \forall p \in P  \\
		    \label{eq:mip3b-personin}
			&\sum\limits_{l \in L_{\phi_p}^-} x_{l} = 1  & \forall p \in P  \\
            \label{eq:mip3b-flowcons1}
			&\sum\limits_{l \in L^{-}_{vp}} x_{l} = \sum\limits_{l \in L^+_{v}} x_{l}  &\forall v \in A_p, p \in P  \\ 
			\label{eq:mip3b-flowcons2}
			&\sum\limits_{l \in L_{v}^-} x_{l} = \sum\limits_{l \in L_{vp}^{+}} x_{l}  &\forall v \in B_p, p \in P \\ 
             \label{eq:mip3b-time-constraints2}
            \tau_l + t_l + &s_v - \tau_n \leq M (2 - x_l - x_n) \quad \forall l \in L^-_{vp}, n \in &L^+_{vp}, v \in V' \cup \{\gamma_p,\phi_p\}, p \in P  \\
			\label{eq:mip3b-taus}
			& \tau_{l} \geq 0	&\forall l \in L 
\end{eqnarray}

Constraints~\eqref{eq:mip3b-subtour-elimination} set the time variables and take care of subtour elimination within trips.
Constraints~\eqref{eq:mip3b-30min} ensure that a user is leaving at the latest $h$ minutes after the end of the task.
Constraints~\eqref{eq:mip3b-planning-horizon-end} restrict the latest departure time at any task to be at the end of the time horizon. 
Constraints~\eqref{eq:mip3b-personout} and \eqref{eq:mip3b-personin} make sure that each user is starting her route in node $\gamma_p$ and ending in node $\phi_p$.
Constraints~\eqref{eq:mip3b-flowcons1} and \eqref{eq:mip3b-flowcons2} balance the flows of start and end nodes of user $p$.
Constraints~\eqref{eq:mip3b-time-constraints2} eliminate parallel trips. 
Finally, constraints~\eqref{eq:mip3b-taus} make sure that decision variables $\tau$ are non-negative.

\subsubsection{Model 4 (\mipvier)}
\label{sec:mip4}

Lastly, in addition to a flexible sequence of tasks, in model \mipvier\, we add time-dependent MOT preferences to the model. This is mainly done by adapting the graph and by adding one constraint {to} the model \mipdreiperson.

We discretize time in intervals of $\alpha$ minutes and duplicate each leg $l \in L$ for each interval. Note that time-dependent MOT preferences are derived from the user preference values $\sigma^t_{pk}$.

We extend the leg information by adding the start and end times of the interval lying on the leg, this results in the tuple $\{(u_l,y_l,z_l,m_l,c_l,\theta_l,o_l,e_l)\}$ where $o_l$ gives the start time and $e_l$ the respective end time of the interval.

Finally, we append the following constraints to model \mipdreiperson: 
\begin{equation}
     o_l x_l \leq \tau_l \leq e_l \quad \forall l \in L 
     \label{eq:addmip4}
\end{equation}

Constraints~\eqref{eq:addmip4} make sure that $\tau_l$ of leg $l$ lies within the predetermined times. 

\noindent
The resulting model relies on both binary and continuous variables. We adapt this and use a re-formulation that is of exponential size but relies on binary variables only. We replace constraints \eqref{eq:mip3b-subtour-elimination}, \eqref{eq:mip3b-30min}, \eqref{eq:mip3b-planning-horizon-end}, \eqref{eq:mip3b-time-constraints2}, \eqref{eq:mip3b-taus}, and \eqref{eq:addmip4} by infeasible path constraints \cite{Ascheuer2000} (for car routes and user routes), and subtour elimination constraints.

Let $\mathcal{R}_{car}$ denote the set of infeasible car routes, and $\mathcal{R}_{p}$ be the set of infeasible user routes. {$V(S)$} gives the {nodes} of the set $S$, where $S$ is a subset of the set of {nodes} on a trip $G_r$. Legs of an infeasible path $\rho$ are denoted as $L(\rho)$.
Model \mipvierb\ can be stated as follows:

\begin{eqnarray}
\mbox{min}	&\eqref{eq:mip1-OF1} \nonumber \\
\mbox{min}	&\eqref{eq:mip1-OF2} \nonumber \\
\mbox{s.t}  &\eqref{eq:mip1-coverevery}-\eqref{eq:mip1-bins},\eqref{eq:mip3b-personout}-\eqref{eq:mip3b-flowcons2} \nonumber \\
    &\sum\limits_{l \in L(\rho)} x_l \leq |L(\rho)| - 1  &\forall \rho \in \mathcal{R}_{car}
     \label{eq:infeasible-path-constraint1} \\
        & \sum\limits_{l \in L(\rho)} x_l \leq |L(\rho)| - 1  &\forall \rho \in \mathcal{R}_p
     \label{eq:infeasible-path-constraint2} \\
         & \sum\limits_{l {\in L(S)}} x_l \leq |{S}| - 1  &\forall S \subseteq {G_r}, r \in R, {S \neq \emptyset}
     \label{eq:subtour}
\end{eqnarray}

Constraints~\eqref{eq:infeasible-path-constraint1}-\eqref{eq:infeasible-path-constraint2} eliminate the infeasible paths of cars and users. We sum over all legs $l$ of the respective infeasible path $\rho$, and set it infeasible by denoting that at least one leg cannot be on the route.
Constraints~\eqref{eq:subtour} are subtour elimination constraints. We set the constraints for all trips $r$ where we store the {nodes} of each trip in the set $G_r$.

\section{Solution approach}
\label{sec:solutionmodel}

In the following, we first introduce valid inequalities in Section~\ref{sec:validineq}. 
By embedding the models into bi-objective optimization frameworks, described in Section~\ref{sec:bioframework}, the scalarized models \mipeins, \mipzwei\ and \mipdreiperson\ are solved with CPLEX. We can solve real-world sized instances within seconds, as we will show in our computational results. However, as expected, \mipvier\ is more challenging to solve. Therefore, we develop a branch-and-cut algorithm in Section~\ref{sec:branchncut} for model \mipvierb. 

\subsection{Valid inequalities}
\label{sec:validineq}

In order to strengthen the models \mipdreiperson, \mipvier, and \mipvierb, the following set of valid inequalities is used. 

\noindent
We know that all legs of a trip must be covered by a single MOT. Therefore, we can say that either MOT $k$ is going into node $v$, or any other MOT $g \neq k$ out of a node $v$, but not both. Assuming that the ingoing legs of a node $v$ are stored in the set $L^-_{vg}$ and all outgoing legs of a node $v$ are stored in the set $L^+_{vk}$, we can state:

\begin{equation}
     \sum_{l \in L^+_{vk}} x_l + \sum_{g \in K: g \neq k}\sum_{l \in L^-_{vg}} x_l = 1 \quad \forall v \in V', k \in K 
     \label{eq:mots}
\end{equation}

\noindent
In \mipdreiperson, \mipvier, and \mipvierb, we only require that the sum over all outgoing legs of a node must be equal to 1. In the following valid inequality the sum over all ingoing legs $l \in L_{vk}^-$ using MOTs $k$ of a node $v$ has to be equal to 1:

\begin{equation}
        \label{eq:allin}  
		\sum\limits_{k \in K} \sum\limits_{l \in L_{vk}^-} x_{l} = 1  \quad \forall v \in V'  
\end{equation}

\noindent
Since a car may cover more than one trip, but has to take at least one if it departs from the depot $d$, the number of trips started with a car (leaving from any node $a \in A$) has to be greater or equal to the sum of cars leaving any depot $d \in D$. Ingoing legs of the start nodes $a$ using MOT $k$ are given in the set $L_{ak}^-$, outgoing legs of the depot $d$ are given in the set $L_{dk}^+$. The constraint is valid for cars only. Thus, we sum over all the ingoing legs of any node $a$, which then has to be greater or equal to the sum over all outgoing legs of any depot $d$:

\begin{equation}
        \label{eq:carstarts}
        \sum\limits_{a \in A}
		\sum\limits_{l \in L_{ak}^-}  x_{l} \geq \sum\limits_{d \in D} \sum\limits_{g \in L_{dk}^+}  x_{g} \quad \textrm{with k = car}
\end{equation}

\noindent
Assuming that a user $p$ has been assigned a single task only, then a full user route will be: $\gamma_p - a_p - q - b_p - \phi_p$. This means, the shortest possible user route consists of four legs. Assuming that all legs assigned to a user $p$ are stored in the set $L_p$, we can formulate:

\begin{equation}
    \label{eq:minlegsperson}
    \sum\limits_{l \in L_p} x_l \geq 4  \quad \forall p \in P
\end{equation}

\noindent
Assuming that a trip has at least one task, then each trip will consist of at least three nodes ($a-q-b$), and thus two legs. The sum over all legs of a trip $r$ is at least the number of nodes assigned to the respective trip, given in the set $G_r$, minus 1. :

 \begin{equation}
    \label{eq:legstrips}
    \sum\limits_{l \in L^-_v: v \in G_r} x_l \geq |G_r| - 1   \quad \forall r \in R
\end{equation}

\noindent
As we know the number of tasks a person is covering, we also know the number of legs the person will cover in the solution. Therefore, we can introduce the following constraint where $L_p$ is the set of legs of a person $p$ and $V_p$ gives the nodes assigned to person $p$:

 \begin{equation}
    \label{eq:nbrpersonlegs}
    \sum\limits_{l \in L_p} x_l = |V_p| - 1   \quad \forall p \in P
\end{equation}

\noindent
We add cycle constraints, meaning that we can only go either from $v$ to $v'$ or from $v'$ to $v$, but not both. We store all legs that start in $v$ and end in $v'$ in the set $L^{(v,v')}$ and vice versa in $L^{(v',v)}$. With this we formulate the following valid inequality:

 \begin{equation}
    \label{eq:circleconstr}
    \sum\limits_{l \in L^{(v,v')}} x_l +  \sum\limits_{l \in L^{(v',v)}} x_l \leq 1 \quad  \forall (v,v') \in L
\end{equation}

The above valid inequalities are used to strengthen \mipdreiperson, \mipvier, and \mipvierb. We now propose additional valid inequalities which are used to strengthen only \mipvier\ and \mipvierb.
\smallskip

\noindent
Let us consider a node $v$, a leg $l$ leaving the node $v$, and an ingoing leg $g$. As described in Section 3.2.4, for the time-dependent setting of the model, the legs contain intervals with the possible start and end time information ($o,e$). With this, we know that the start and end times of the outgoing leg $l$ has to be greater than the times of the ingoing leg $g$. Therefore, if the start and end times of the ingoing leg $g$ are greater than the times of the outgoing leg $l$, meaning that the ingoing leg would happen later in time, only one of them can be used:

\begin{equation}
     o_l < o_g {\wedge} e_l < e_g \iff x_l + x_g \leq 1 \quad \forall l \in L^+_v, g \in L^-_v, v \in V'
     \label{eq:risinginterval}
\end{equation}

\noindent
As any outgoing leg of a node $v$ has to be later than the ingoing leg of the respective node, we can further eliminate all outgoing legs of a node $v$ that are timed before a chosen ingoing leg of the respective node.
Therefore, we adapt equation~\eqref{eq:risinginterval}, where we assume an ingoing leg $g \in L^-_v$ of a node $v$ and sum over all outgoing legs $l \in L^+_v$ with {smaller start and end times} as the ingoing leg, thus $o_l < o_g$ and $e_l < e_g$. Then at most one of the respective legs can be chosen. Conversely, considering an outgoing leg $l$ and summing over all ingoing legs $g \in L^-_v$ with an interval greater than the one of the outgoing leg ($o_g > o_l$,$e_g > e_l$), we can again say that at most one leg can be chosen. Both valid inequalities can be formulated as follows:

\begin{eqnarray}
     x_g + \sum_{l \in L^+_v: o_l < o_g {\wedge} e_l < e_g} x_l \leq 1 \quad \forall g \in L^-_v, v \in V'
     \label{eq:vi-time1} \\
     x_l + \sum_{g \in L^-_v: o_g > o_l {\wedge} e_g > e_l} x_g \leq 1 \quad \forall l \in L^+_v, v \in V'
     \label{eq:vi-time2}
\end{eqnarray}

\noindent
If the beginning of the interval $o_l$ of the outgoing leg $l$ is greater than the end of the interval $e_g$ of the ingoing leg $g$ plus the time of the ingoing leg $t_g$ plus the service time at the node $s_v$ plus the maximum waiting time $h$, then these legs are not compatible in time. Again, considering a node $v$ with outgoing legs $L^+_v$ and ingoing legs $L^-_v$, we can state the following valid inequalities:

\begin{eqnarray}
     x_g + \sum_{l \in L^+_v: o_l > e_g+t_g+s_v+h} x_l \leq 1 \quad \forall g \in L^-_v, v \in V'
     \label{eq:vi-time1} \\
     x_l + \sum_{g \in L^-_v: o_l > e_g+t_g+s_v+h} x_g \leq 1 \quad \forall l \in L^+_v, v \in V'
     \label{eq:vi-time2}
\end{eqnarray}

\noindent
If the beginning interval $o_g$ of the ingoing leg $g$ plus the travel time of the ingoing leg $t_g$ plus the service time of the node $s_v$ is greater than the end of the interval $e_l$ of the outgoing leg $l$, then these legs cannot be used together. We can again put this into two valid inequaities as follows:

\begin{eqnarray}
     x_g + \sum_{l \in L^+_v: o_g+t_g+s_v > e_l} x_l \leq 1 \quad \forall g \in L^-_v, v \in V'
     \label{eq:vi-time1} \\
     x_l + \sum_{g \in L^-_v: o_g+t_g+s_v > e_l} x_g \leq 1 \quad \forall l \in L^+_v, v \in V'
     \label{eq:vi-time2}
\end{eqnarray}

\subsection{Branch-and-cut for \mipvierb}
\label{sec:branchncut}

In order to solve model \mipvierb, we develop a branch-and-cut algorithm. 
Branch-and-cut algorithms make use of a subset of constraints and iteratively add further constraints in a cutting-plane fashion. Usually, constraint sets of exponential size are excluded which reduces the model to a reasonable size. In our case, we separate the infeasible path constraints~\eqref{eq:infeasible-path-constraint1}-\eqref{eq:infeasible-path-constraint2} but we enumerate all subtour elimination constraints, since trips are usually very short. Separation algorithms are then called to determine whether the current solution is feasible by checking the omitted constraints. Note that the separation algorithms can be called on any relaxed solution or only on incumbent ones. Our strategy is based on the latter case, where we only call the algorithms if a new incumbent solution is found. If the separation algorithm detects a violation, the respective constraint is added as a cut to the model and the model is consecutively resolved. This is repeated until no violating constraints are detected and an optimal solution is found. 

In our model a route (path) may be infeasible due to (i) user related constraints, (ii) shared cars related constraints, and (iii) synchronization requirements between user and car routes. Therefore, we first check if all user routes are feasible, then if all car routes are feasible and finally if they are both simultaneously feasible. The respective separation procedures are described in the following. 

\subsubsection{Separation of infeasible user routes}

We separate infeasible user routes for each user $p \in P$. Let $\overline{x}$ denote the solution at the current node in the branch-and-bound tree. We start the construction of the route $\rho$ at node $\gamma_p$. We denote the currently considered node as node $v$. From the starting point, we append the outgoing leg $l$ at node $v$ ($v$.outgoing) if $\overline{x}_l=1$ to the route $\rho$, and update $v$ to be the end node of leg $l$ ({$l$}.endNode). We do this until we hit the user end depot $\phi_p$. 

In the following, we consider a forward slack $F$, consisting of an accumulated waiting time $\mathcal{W}$ and a value stating how much we could move the whole route such that the solution would still be feasible, given as $\Delta$, and $F=\mathcal{W}+\Delta$. The current time stamp is given as $\tau$.
Before checking the route $\rho$ for time feasibility, we initialize $F$, $\mathcal{W}$, $\tau$ to $0$, and $\Delta=\infty$.

We iterate through the route as long as all time constraints are respected. We start by checking the second leg $l$ on route $\rho$ and systematically take the consecutive one. Thus, considering the current leg $l$ leaving node $v$, we set $\tau=\tau+s_v+t_{l-1}$ and update $\mathcal{W}$ and $\Delta$. The accumulated waiting time is calculated as the current waiting time plus either the maximum possible waiting time $h$ or the remaining time to the end of the interval $e_l$, such that $\mathcal{W}=\mathcal{W}+min\{max\{0,e_l-\tau\},h\}$. We can further push the whole route to the end of the given interval $e_l$ or by the previously stored $\Delta$. We update $\Delta=min\{\Delta,max\{0,e_l-(\tau+\mathcal{W})\}\}$ and compose $F=\mathcal{W}+\Delta$.
If the current time $\tau$ lies within the respective interval of leg $l$ ($o_l,e_l$), we can proceed to the next leg. If not, we try to push the route to the starting interval $o_l$ of the current leg $l$, but at maximum by adding $F$, such that $\tau=\tau+min\{max\{0,o_l-\tau\},F\}$.
If the adapted $\tau$ violates the timing restrictions, the corresponding infeasible path constraint is generated.
If $\tau$ is feasible ($o_l \leq \tau < e_l$), we can update $\mathcal{W}$ and $\Delta$, and proceed with the next leg. To update the values, we have to deduct the respective time used up of the forward slack. For this, we first adapt $\Delta$ by stating $\Delta=\Delta+min\{\mathcal{W}-(\tau-\tau'),0\}$, where $\tau'$ denotes the time stamp before adding the time slack $F$. The waiting time is updated as $\mathcal{W}=max\{\mathcal{W}-(\tau-\tau'),0\}$. The pseudocode is outlined in the Appendix in Algorithm~\ref{algo:user}.

\subsubsection{Separation of infeasible car routes}

We further aim to detect infeasible paths regarding cars violating time constraints. 
We adopt the same idea as above, except following car routes. Starting depots of cars are $d \in D$. Note that, as we might have more than one trip originating from one node $d$, we slightly adapt the construction of the route $\rho$ by considering node $d$ multiple times as a starting node for the construction of the route $\rho$. We store the outgoing leg $l$ of node $v$ with $\overline{x}_l=1$ and the MOT $k=car$ in the route $\rho$. While constructing, we save the number of trips on the current route, as we only consider routes with more than one trip. Timing restrictions for a single trip are already covered in the user route separation. If the route $\rho$ consists of multiple trips, we follow the same steps as previously described in the separation algorithm of user routes. The pseudocode is given in Algorithm~\ref{algo:vehicle} in the Appendix.

\subsubsection{Synchronization of routes}

It is not sufficient to check user and car routes separately for infeasibility.
We also have to check if the user and car routes are synchronized, i.e. if the user who has taken over a car is at the depot at the respective time. In order to do so, we consider the whole solution and we store the used legs in the subset $L'$, and obtain the sets $L'^-_{vk}, L'^+_{vk}, L'^-_{vp}, L'^+_{vp}$ ($\overline{x}_l$ = 1 in the current solution), and we solve the following small LP derived from constraints~\eqref{eq:mip3b-subtour-elimination}, \eqref{eq:mip3b-30min} and \eqref{eq:mip3b-time-constraints2}:
\begin{eqnarray}
        \label{eq:cuts1}
		&\tau_l + t_l + s_v \leq \tau_{{n}}  &\forall l \in L'^-_{vk}, n \in L'^+_{vk}, v \in V', k \in K \\
		\label{eq:cuts3}
        &\tau_l + t_l + s_{{v}}  \leq \tau_{{n}} &\forall l \in L'^-_{vp}, n \in L'^+_{vp}, v \in V' \cup U, p \in P \\
        \label{eq:cuts2}
        &\tau_{{l}} + t_{{l}} + s_v \geq \tau_{{n}} - h  &\forall l \in L'^-_{vk}, n \in L'^+_{vk}, v \in V', {k \in K}
\end{eqnarray}

Constraints~\eqref{eq:cuts1} and \eqref{eq:cuts3} synchronize the car and user route with the decision variable $\tau$. Furthermore constraints~\eqref{eq:cuts2} make sure that the waiting time at a node is not exceeded. 

The above constraints are infeasible if the respective user and car are not at the same time at the same place. Therefore we can assume that either a car trip or an arc connecting the user trips is infeasible. Thus, we insert into the set $L''$ all legs from the set $L'$ that are taken by car or are connecting user trips in the current incumbent solution, and we add the following constraint:

\begin{equation}
		\label{eq:cutoff-synch}
        \sum_{l \in L''} x_l \leq |L''| - 1 
\end{equation}

\subsubsection{Strengthened infeasible path constraints}
\label{sec:cutsadding}

The infeasible paths introduced before in the form of constraint~\eqref{eq:infeasible-path-constraint1} and constraint~\eqref{eq:infeasible-path-constraint2} are very weak. We strengthen them as follows:

\begin{equation}
		\label{eq:infeasiblepath-strengthened}
        \sum_{l \in \rho} x_l + \sum_{l \in \mathcal{L}'} x_l +  \sum_{l \in \mathcal{L}''} x_l  \leq |\rho| - 1 \\
\end{equation}

Let $\mathcal{L}'$ contain all legs with the same start node $y$, end node $z$ and MOT $k$ but earlier or later intervals $(o,e)$ than the last checked leg of the separation algorithm, i.e., where the infeasiblity was detected. Let $l'$ be the last checked leg, and $\tau$ the current departure time. If $\tau > e_{l'}$, meaning that we have jumped over the interval, then the set $\mathcal{L}'$ contains all legs with the same respective $y,z,k$ but $o < o_{l'}$. This means that if we missed the interval, then also all prior ones will be too early. Conversely, if the interval of leg $l'$ could not be reached, thus $\tau < o_{l'}$, we put all legs with the same $y,z,k$ as leg $l'$ but $o > o_{l'}$ into the set $\mathcal{L}'$. Hence, if we were not able to reach the respective interval, then also all later legs will not be reachable.

The set $\mathcal{L}''$ also depends on whether we are not able to reach the leg's interval or we miss it. We consider all legs on the route $\rho$ except the last checked leg $l'$, denoted as $\rho'$.  
Considering $\tau < o_{l'}$, thus the time stamp lies before the start of the interval, then the set $\mathcal{L}''$ contains the respective counterparts of all legs in $\rho'$  with the same $y,z,k$ but with an interval that lies behind the last saved $\tau$.
If we miss the interval of $l'$, such that $\tau > e_{l'}$, we assume that we cannot push any prior leg any further. In this case, we detect the respective duplications of the legs in $\rho'$ with a higher interval such that the interval of any leg $l$ is greater than $o_{l''}$, where $l''$ depicts the leg assigned to $\rho$.

Moreover, we store all checked legs to the vector $\rho_{short}$. We know that the last leg is incompatible with the prior ones, and can therefore add the following constraint:

\begin{equation}
		\label{eq:infeasiblepath-strengthened1}
        \sum_{l \in \rho_{short}} x_l + \sum_{l \in \mathcal{L}'} x_l +  \sum_{l \in \mathcal{L}''} x_l  \leq |\rho_{short}| - 1 
\end{equation}

\subsection{Bi-objective frameworks}
\label{sec:bioframework}

We embed our models into two bi-objective frameworks. For \mipeins, \mipzwei, and \mipdreiperson\ we use the $\epsilon$-constraint method. The branch-and-cut algorithm to solve \mipvierb\ is embedded into both frameworks, namely the $\epsilon$-constraint method and a weighting binary search method. 

\subsubsection{The $\epsilon$-constraint method}
The $\epsilon$-constraint method iteratively solves single-objective problems where one objective is kept in the objective function and the other one is moved to the set of constraints. After each iteration the respective constraint in the constraint set is reduced by a certain $\epsilon$. As we only consider integer variables and coefficients we define the $\epsilon$-value to be $1$. For example, let us consider the cost function~\eqref{eq:mip1-OF1} as the main objective, and the preferences objective~\eqref{eq:mip1-OF2} is moved to the constraint as: $\sum_{l \in L} \theta_l x_l \leq \Omega - \epsilon$. $\Omega$ is iteratively adapted, inserting the preference value from the previous subproblem and is initially set to $\infty$. We solve the problems in a lexicographic order, meaning that in each iteration two MIPs are solved. The algorithm stops once the second extreme point of the Pareto frontier (with the minimal second objective) is reached.

\subsubsection{A weighting binary search method}
As a second framework we use a binary search in the objective space that is based on the algorithm introduced by \citet{Riera2005}. The idea is to use a weighting method and iteratively enumerate the Pareto frontier. To start the algorithm the extreme points of the Pareto frontier are calculated and stored as $(f^{(1)}_1,f^{(1)}_2)$ and $(f^{(2)}_1,f^{(2)}_2)$. $f^{(1)}_1$ and $f^{(2)}_1$ give the first, e.g. cost, solutions, of the respective extreme points and conversely $f^{(1)}_2$ and $f^{(2)}_2$ give the value of the second objectives, in our case: preferences. Thus, the objective value is set as $\omega f^*_1 + (1-\omega)f^*_2$, where $f^*$ denotes the cost and preference function of the new solution. The weight $\omega$ is calculated as $\frac{\alpha}{\alpha + 1}$, where $\alpha = \frac{f^{2}_2 - f^{1}_2}{f^{1}_1 - f^{2}_1}$. At each iteration we add three constraints: (1) $f^*_1 < f^{2}_1$, (2) $f^*_2 < f^{1}_2$, and (3) $\omega f^*_1 + (1 - \omega)f^*_2 \leq \omega f^{(1)}_1 + (1 - \omega) f^{(2)}_2 - 1$. The latter one makes sure that only non-dominated points are generated. The values of the new solutions are then used for the following iterations where the next weights will be calculated with the values $(f^{(1)}_1,f^{(1)}_2)$ and $(f^{*}_1,f^{*}_2)$, as well as with $(f^{*}_1,f^{*}_2)$ and $(f^{(2)}_1,f^{(2)}_2)$. The algorithm terminates once no more values can be taken to calculate new weights.

\paragraph{Enhancements}
For both methods we seize the bi-objective characteristics of our problem: we store the cuts generated in the prior iterations and add them as constraints to the next models. Considering the $\epsilon$-constraint method, we do this within one iteration (the min cost problem receives the cuts from the min preference model), as well as from one iteration to another. As for the binary search, we only solve one MIP with the respective objective function within each iteration. Therefore, we only pass on cuts from one solution to another.

\section{Computational study} \label{sec:comp}

The models and the branch-and-cut algorithm are implemented in C++ and solved with CPLEX 12.9. Tests are carried out using one core of an Intel Xeon Processor E5-2670 v2 machine with 2.50 GHz running Linux CentOS 6.5. Unless otherwise stated a time limit of 12 hours is used. 
\subsection{Test instances}\label{sec:data}

For our computational study we use realistic benchmark instances based on available demographic, spatial and economic data of Vienna, Austria. They are based on those used in \citet{enzi2020modeling} and \citet{Enzi2020}. Note that the instances represent a company within a city, thus the data does not aim to replicate the population of the whole city.

One instance set represents a distinct company consisting of one or more offices (or depots) $D$ and users, i.e. employees, $P$. The number of tasks and their location are randomly generated. 

In the original instances, each user may use a subset of the available MOTs $K^p \subseteq K$. Based on this binary assignment of MOTs to users, we generate preference scores on a scale from 1-10, where 1 is best and 10 is worst. For example, if a user has cars in her set of MOTs but no public transportation, then this user will get a lower (better) score on cars and a higher (worse) one on public transportation. The detailed assignments used for the following study is included in Table~\ref{tab:preferencescore-base} in the Appendix. 
In the time-dependent setting, we consider seven different time periods $t$: pre-rush-hour, rush-hour, after rush-hour, normal day-time traffic, pre-rush-hour, rush-hour and after rush-hour. Here we deduct/add for each preference score a certain number (see Table~\ref{tab:preferencescore-period} in the Appendix). 
Furthermore, we implement an increase/decrease in cost and time for the respective time periods (see Table~\ref{tab:costtime-factor} in the Appendix). For this we assume a factor $\beta$ which is then multiplied with the base cost. For example, we assume that taking the car during rush-hour takes longer than at noon. We assume $\beta = 1.4$, which is then multiplied with the base cost, e.g. 5. This gives us cost of 7 for the rush-hour for the respective leg. Naturally also the driving times of the legs are adapted accordingly. We calculate a weighted average of cost and preferences if a leg covers more than one periods. 

Three different MOTs are considered: car, public transportation including walk, and bike. For our study we assume that all MOTs have an unrestricted capacity. Note that the original setting assumes a limited and fixed pool of cars, which is reasonable for the discussed problem. However, for our first results for the BiO-MMCP we decided to let the number of cars be unlimited, to explore the computational efficiency without restricting the number of shared cars.
Distances, time and cost are calculated between all nodes for all modes of transport. 
Emissions are translated into costs and, together with variable distance cost and cost of time, included into the overall cost calculations and summarized in $c_l$. The respective preference value $\theta_l$ is taken from the above presented values.

Instances are named as \texttt{E\_}$|P|$\texttt{\_}$I$, where $|P|$ is the number of users, and the instance number $I$ is between 0 and 9. For example, the first instance in the set of instances with 20 users ($|P| = 20$) is denoted \texttt{E\_20\_0}. For instance group with $|P|$ users, we solve a set of 10 instances (\texttt{E\_u\_0} to \texttt{E\_u\_9}) and report the average values.

\subsection{Enhancements and preprocessing} 
\label{sec:enhance-prep}

In the following paragraphs we shortly list the enhancements and preprocessing that we conducted. 

\paragraph{Relative MIP-gap}

In our first tests, CPLEX provided weakly dominated solutions or skipped some of the solutions from the Pareto set due to the default relative MIP-gap. Therefore we put a strict MIP-gap tolerance of 0.0000. We compared the output with different tolerances regarding computational efficiency and could not notice a remarkable difference. Therefore, unless otherwise stated, the computational results are based on a MIP-gap tolerance of 0.0000.

\paragraph{Warm start}

For models \mipdreiperson\ and \mipvier, we provide CPLEX with a starting solution. The starting solution is constructed by simply reading the sequence of the tasks as given in the instance file. For model \mipvier, we also track the according times and make sure that the times and intervals match. In the starting solution, public transportation is used on all trips. Moreover, after each iteration we store the solution and provide CPLEX with a MIP start. The MIP start will be infeasible but values can be stored for a possible repair.

\paragraph{Graph reduction}

Initially for model \mipvier, we duplicate each leg every $\alpha$ minutes. Assuming that we have a planning horizon of 12 hours and discretize time in steps of 15 minutes, we end up with 48 duplicates of one leg. However, these legs are very similar to each other or even equal as the time periods $t$ may cover various hours. Therefore, in order to reduce the size of the graph, we merge legs with equal weights. 

Table~\ref{tab:basic-info} gives an overview of the size of the graphs. The table gives information on the introduced models for an increasing number of users ($|P|$ = 20, 50, 100, 150, 200, 250, 300). For \mipeins\ and \mipzwei\ the underlying graph has the same size as only the preference and cost values on the legs are changing. Row '$|\overline{V'}|$' gives the average number of nodes, '$|\overline{R}|$' the average number of trips, and row '$|\overline{L}|$' the average number of legs in the respective graphs. We observe that the underlying graphs of the first two models have a moderate number of legs as the sequences are predetermined. In models \mipdreiperson, \mipvier, and \mipvierb\ the sequence is subject of determination which leads to an increasing number of connecting legs, which is increased even further when time-dependency is considered in models \mipvier\ and \mipvierb. Row '\mipvier,\mipvierb+GR' shows the number of the legs in the graph after the graph reduction.

\begin{table}
  \caption{Average number of nodes ($|\overline{V'}|$), trips ($|\overline{R}|$) and legs ($|\overline{L}|$) for models \mipeins, \mipzwei, \mipdreiperson, \mipvier, \mipvierb\ and an increasing number of users $|P|=20,50,100,150,200,250,300$. Row '\mipvier,\mipvierb+GR' gives the average number of legs after the graph reduction.}
   \label{tab:basic-info}%
  \small
    \begin{tabular}{llrrrrrrr}
    \hline\noalign{\smallskip}
          & $|P|=$  & 20    & 50    & 100   & 150   & 200   & 250   & 300 \\
    \noalign{\smallskip}\hline\noalign{\smallskip}
    $|\overline{V'}|$ &       &                    95  &                  242  &                  476  &                  714  &                  951  &                 1,179  &                 1,422  \\
    $|\overline{R}|$ &       &                    31  &                    76  &                  147  &                  218  &                  287  &                    358  &                    427  \\
    \noalign{\smallskip}\hline\noalign{\smallskip}
     \multirow{3}[2]{*}{$|\overline{L}|$} & \mipeins, \mipzwei &                  416  &              1,188  &              2,612  &              4,340  &              6,315  &                 8,734  &              11,038  \\
          & \mipdreiperson  &                  947  &              4,190  &            13,394  &            27,756  &            46,503  &              70,492  &              99,462  \\
        &      \mipvier,\mipvierb &            13,479  &            41,077  &            93,079  &          150,877  &          216,989  &         276,626     &            356,713  \\
          & \mipvier,\mipvierb+GR  &              3,984  &            13,995  &            34,780  &            61,310  &            92,790  &            127,155  &            167,645  \\
   \noalign{\smallskip}\hline
    \end{tabular}%
\end{table}%

\subsection{Algorithmic tests}

In this section we study the computational efficiency of the introduced variants of the model. We start by comparing the four models (\mipeins, \mipzwei, \mipdreiperson, \mipvier) in their basic form, i.e., without adding valid inequalities or using the branch-and-cut algorithm. Afterwards we analyze the impact of valid inequalities on model \mipdreiperson. Finally, we focus on solving the most challenging model \mipvier. We compare the reformulation of \mipvier\ to \mipvierb, i.e., if the branch-and-cut algorithm comes with any improvements in computational efficiency. We aim to detect the enhancements by adding valid inequalities, using the branch-and-cut algorithm and choose the best framework (out of the two introduced) to solve model \mipvierb.

\subsubsection{Comparison of models \mipeins, \mipzwei, \mipdreiperson, and \mipvier}

In a first step, we compare the four models regarding their run times. Table~\ref{tab:mips-basic} shows the average computational time in seconds needed to solve an instance group.  We first look into results without any valid inequalities or cut generation, given in the rows \mipeins, \mipzwei, \mipdreiperson, and \mipvier. The models are embedded in the $\epsilon$-constraint method and enumerated by setting either the cost function as the objective (\costfirst) or the user preferences as the objective (\preffirst). Results are given for instance sets for which we were able to solve all 10 instances. Run times for \mipeins\ and \mipzwei\ are very short. We can solve real-world sized instances with 300 users in less than 5 minutes. For \mipeins\ the direction of the $\epsilon$-constraint method has no impact. In the case of \mipzwei\ setting \preffirst\ as first objective results in shorter run times. Models \mipdreiperson\ and \mipvier\ are 'harder' to solve. 
For model \mipdreiperson\ we can see a significant increase in the average run times for the instance group with $|P|=50$. The largest instance set we can solve comprises 100 users in the case of \mipdreiperson. Adding valid inequalities reduces computational times by a factor of 3 for this instance size (\mipdreiperson-VI) and $|P|=100$. 
With \mipvier\ we cannot solve any complete instance set. We will go into more detail on \mipvier, its possible extensions and the respective results later. Using the best setting of the proposed branch-and-cut based algorithm, we are able to enumerate the whole Pareto frontier within about 3,000 seconds, on average. 

\begin{table}
  \caption{Average computational times in seconds for models \mipeins, \mipzwei, \mipdreiperson, \mipdreiperson-VI, \mipvier, \mipvierVIBnCbio and an increasing number of users $|P|=20,50,100,150,200,250,300$ for both directions (\costfirst, \preffirst) in the $\epsilon$-constraint method. \mipdreiperson-VI gives results for the respective model with valid inequalities. \mipvierVIBnCbio shows results for the model \mipvierb\ solved by branch-and-cut and passing cuts to subsequent iterations.}\label{tab:mips-basic}%
  \small
    \begin{tabular}{clrcccccc}
    \hline\noalign{\smallskip}
          & $|P|=$  & 20    & \multicolumn{1}{r}{50} & \multicolumn{1}{r}{100} & \multicolumn{1}{r}{150} & \multicolumn{1}{r}{200} & \multicolumn{1}{r}{250} & \multicolumn{1}{r}{300} \\
   \noalign{\smallskip}\hline\noalign{\smallskip}
    \multirow{2}[2]{*}{\mipeins} & \costfirst &                   0.3  & \multicolumn{1}{r}{                  3.0 } & \multicolumn{1}{r}{               15.8 } & \multicolumn{1}{r}{               45.2 } & \multicolumn{1}{r}{               88.1 } & \multicolumn{1}{r}{             163.4 } & \multicolumn{1}{r}{             247.3 } \\
          & \preffirst &                   0.3  & \multicolumn{1}{r}{                  3.0 } & \multicolumn{1}{r}{               15.8 } & \multicolumn{1}{r}{               43.6 } & \multicolumn{1}{r}{               85.1 } & \multicolumn{1}{r}{             160.9 } & \multicolumn{1}{r}{             238.5 } \\
    \noalign{\smallskip}\hline\noalign{\smallskip}
    \multirow{2}[2]{*}{\mipzwei} & \costfirst &                   0.7  & \multicolumn{1}{r}{                  5.5 } & \multicolumn{1}{r}{               35.4 } & \multicolumn{1}{r}{               92.0 } & \multicolumn{1}{r}{             188.2 } & \multicolumn{1}{r}{             319.8 } & \multicolumn{1}{r}{             428.5 } \\
          & \preffirst &                   0.5  & \multicolumn{1}{r}{                  4.4 } & \multicolumn{1}{r}{               29.8 } & \multicolumn{1}{r}{               79.8 } & \multicolumn{1}{r}{             165.3 } & \multicolumn{1}{r}{             288.2 } & \multicolumn{1}{r}{             399.9 } \\
    \noalign{\smallskip}\hline\noalign{\smallskip}
    \multirow{2}[2]{*}{\mipdreiperson} & \costfirst &                   7.4  & \multicolumn{1}{r}{             498.5 } & \multicolumn{1}{r}{       17,707.6 } &  -    &  -    &  -    &  -  \\
          & \preffirst &                   7.7  & \multicolumn{1}{r}{         1,037.3 } &  -    &  -    &  -    &  -    &  -  \\
    \noalign{\smallskip}\hline\noalign{\smallskip}
    \multirow{2}[2]{*}{\mipdreiperson-VI} & \costfirst & \multicolumn{1}{r}{ 8.0 } & \multicolumn{1}{r}{ 480.4 } & \multicolumn{1}{r}{         5,743.2 } &  -    &  -    &  -    &  -  \\
          & \preffirst & \multicolumn{1}{r}{ 8.5 } & \multicolumn{1}{r}{ 874.5 } &      \multicolumn{1}{r}{     5,577.5}  &  -    &  -    &  -    &  -  \\
    \noalign{\smallskip}\hline\noalign{\smallskip}
    \multirow{2}[2]{*}{\mipvier} & \costfirst & \multicolumn{1}{c}{-} & -     & -     & -     & -     & -     & - \\
          & \preffirst & \multicolumn{1}{c}{-} & -     & -     & -     & -     & -     & - \\
    \noalign{\smallskip}\hline\noalign{\smallskip}
    \multicolumn{1}{c}{\multirow{2}[2]{*}{\mipvierVIBnCbio}} & \costfirst &          3,511.9  &  -    &  -    &  -    &  -    &  -    &  -  \\
          & \preffirst &          2,730.7  &  -    &  -    &  -    &  -    &  -    &  -  \\
    \noalign{\smallskip}\hline
    \end{tabular}%
\end{table}%

Table~\ref{tab:paretp-sol} summarizes the average number of Pareto optimal solutions per instance set. The number of solutions is moderately increasing with number of users $|P|$. Comparing \mipeins\ and \mipdreiperson\ we see almost the same number of Pareto optimal solutions on average per instance set. If we compare the increased cost in computational complexity coming with \mipdreiperson, we could argue that dissolving the sequences where no time-dependent information is given, is not worthwile. We will investigate the shape of the Parento frontiers in a subsequent section in order to obtain a better understaning of the resulting solutions.
Comparing \mipeins\ with \mipzwei\ we can see a distinct increase of optimal solutions on the frontier, even though we only introduced time-dependent cost and preferences. Finally, \mipvier\ gives by far the highest number of optimal solutions for the small instance set of $|P|=20$. 

\begin{table}
  \caption{Average number of Pareto optimal solutions for models \mipeins, \mipzwei, \mipdreiperson, \mipvier\ for an increasing number of users $|P|=20,50,100,150,200,250,300$.}  \label{tab:paretp-sol}%
  \small
    \begin{tabular}{lrrrrrrrr}
    \hline\noalign{\smallskip}
   \multicolumn{1}{l}{$|P|=$} & 20    & 50    & 100   & 150   & 200   & 250   & 300 \\
    \noalign{\smallskip}\hline\noalign{\smallskip}
    \mipeins  &                       18  &                    73  &                  159  &                  250  &                  349  &                  464  &                  518  \\
    \mipzwei  &                      34  &                  164  &                  346  &                  555  &                  767  &                  993  &              1,073  \\
    \mipdreiperson  &                       18  &                    72  &                  158  &  -    &  -    &  -    &  -  \\
    \mipvier      & 141   &  -    &  -    &  -    &  -    &  -    &  -  \\
    \noalign{\smallskip}\hline
    \end{tabular}%
\end{table}%

\subsubsection{Introducing valid inequalities for model \mipdreiperson}

We now analyze the impact of the proposed valid inequalities (VI) in more detail.
Table~\ref{tab:mip3-visec} presents the computational times in seconds solving \mipdreiperson\ without additional information (\mipdreiperson) and by adding valid inequalities \eqref{eq:mots} - \eqref{eq:circleconstr} as well as subtour elimination constraints~\eqref{eq:subtour} as user cuts (\mipdreiperson-VI). We use both, costs (\costfirst) and preferences (\preffirst) respectively as the 'main' objective function in the $\epsilon$-constraint method. Results are given for $|P| = 100,150$ and listed for each instance. Row '\# solved' shows the number of instances solved with the respective model.
We can observe that for some of the instances, e.g. E\_100\_8, for both \costfirst\ and \preffirst, the execution time is higher with the valid inequalities than without them. However, on average adding additional information in the form of valid inequalities improves computation times by a factor of approximately 4. Even for instance E\_100\_2, where we were not able to enumerate the whole Pareto frontier within 12 hours with the base model, we are now able to get the frontiers from either side in less than 3 hours.  For the case where $|P|=150$ and \preffirst\ we are able to solve all but two instances, however all with relatively long computational times. Direction \costfirst\ shows longer run times for all solved instances, whereas for two more cases, in total four, the total Pareto frontier cannot be enumerated. None of the instances with $|P|=150$ has been solved without the valid inequalities. Furthermore, we were not able to solve any of the instances with $|P|=200$ using \mipdreiperson\ or \mipdreiperson-VI.

\begin{table}
  \caption{Average computational times in seconds for $|P|=100,150$ solving \mipdreiperson\ without valid inequalities (\mipdreiperson) and the same model including valid inequalities (\mipdreiperson-VI) and having either cost (\costfirst) or preferences (\preffirst) set as the objective function in the $\epsilon$-constraint method. '\# solved' shows the number of instances solved. TO = time out.}  \label{tab:mip3-visec}%
  \small
  \setlength{\tabcolsep}{4pt}
    \begin{tabular}{lrrrrrr|lrrrrr}
    \hline\noalign{\smallskip}
      & \multicolumn{2}{c}{\costfirst} &       & \multicolumn{2}{c}{\preffirst} &       &       & \multicolumn{2}{c}{\costfirst} &       & \multicolumn{2}{c}{\preffirst} \\
          &   \mipdreiperson    & \multicolumn{1}{l}{\mipdreiperson-VI} &       &    \mipdreiperson   & \multicolumn{1}{l}{\mipdreiperson-VI} &       &       &   \mipdreiperson    & \multicolumn{1}{l}{\mipdreiperson-VI} &       &    \mipdreiperson   & \multicolumn{1}{l}{\mipdreiperson-VI} \\
\noalign{\smallskip}\hline\noalign{\smallskip}   E\_100\_0 & \textbf{     2,419} &    2,852  &       &      3,212  &    2,628  &       & E\_150\_0 &  TO   &  TO   &       &  TO   &  TO  \\
    E\_100\_1 &      3,329  &    3,104  &       & \textbf{     2,513} &    2,777  &       & E\_150\_1 &  TO   &  TO   &       &  TO   & \textbf{   39,507} \\
    E\_100\_2 &    35,758  & \textbf{   7,275} &       &  TO    &    8,504  &       & E\_150\_2 &  TO   &  TO   &       &  TO   & \textbf{   41,675} \\
    E\_100\_3 &      3,382  &    3,544  &       & \textbf{     3,281} &    3,892  &       & E\_150\_3 &  TO   &    37,060  &       &  TO   & \textbf{   20,507} \\
    E\_100\_4 &    24,162  & \textbf{   4,177} &       &    25,159  &    4,553  &       & E\_150\_4 &  TO   &  TO   &       &  TO   &  TO  \\
    E\_100\_5 &    27,186  &    6,655  &       &    25,587  & \textbf{   6,062} &       & E\_150\_5 &  TO   &    19,766  &       &  TO   & \textbf{   14,098} \\
    E\_100\_6 &    11,152  &    9,814  &       &    11,403  & \textbf{   9,092} &       & E\_150\_6 &  TO   &    18,352  &       &  TO   & \textbf{   13,640} \\
    E\_100\_7 &    24,112  &    9,245  &       &    22,085  & \textbf{   6,880} &       & E\_150\_7 &  TO   &    38,193  &       &  TO   & \textbf{   21,112} \\
    E\_100\_8 &      3,398  &    5,204  &       & \textbf{     2,685} &    3,678  &       & E\_150\_8 &  TO   &    30,083  &       &  TO   & \textbf{   23,643} \\
    E\_100\_9 &    22,771  & \textbf{   5,561} &       &    28,090  &    7,708  &       & E\_150\_9 &  TO   &  TO   &       &  TO   & \textbf{   35,617} \\
\noalign{\smallskip}\hline\noalign{\smallskip}
\# solved & 10 & 10 & & 9 & 10 & &  & 0 & 5 & & 0 & 8 \\
\noalign{\smallskip}\hline
\end{tabular}%
\end{table}%

\subsubsection{Solving model \mipvier}

We now compare the different approaches for solving \mipvier. Table~\ref{tab:mip4-comp} shows the run times for (i) model \mipvier\ (\mipvier($\epsilon$)),  (ii) \mipvier\ with valid inequalities (\mipvierVI($\epsilon$)), (iii) model \mipvierb\ with valid inequalities, and infeasible path constraints in the form of \eqref{eq:infeasiblepath-strengthened}-\eqref{eq:infeasiblepath-strengthened1} added through cut generation and embedded into the $\epsilon$-constraint method (\mipvierVIBnC ($\epsilon$)), (iv) the bi-objective branch-and-cut, which is similar to the previous variant but we pass the cuts generated as constraints from one solution to another (\mipvierVIBnCbio($\epsilon$)), (v) model \mipvierb\ solved by branch-and-cut embedded in the weighting binary search method (\mipvierVIBnC ($\omega$)), and (vi) the branch-and-cut used to solve \mipvierb\ using the weighting binary search method and passing cuts to subsequent iterations (\mipvierVIBnCbio ($\omega$)). Again all results are given for both directions, \costfirst\ and \preffirst\ in the case of the $\epsilon$-constraint scheme. In the case of the binary search, both objectives are combined in one weighting objective function. Times are in seconds. Row '\# solved' gives the number of instances solved. Results are given for each instance for $|P|=20$. 

Using model \mipvier ($\epsilon$) and the direction \costfirst, only one instance is solved, using \preffirst\ as the main objective, two instances can be solved within 12 hours of computation time. Adding valid inequalities (\mipvierVI($\epsilon$)), we are able to increase the number of instances solved to 6 for the direction \costfirst\ and to 7 for the direction \preffirst. Still for most of the instances the run times exceed 10,000 seconds. 

Moving from the model with the time variables (\mipvier) to the entirely integer model (\mipvierb) with cut generation, we can improve run times considerably by at least a factor of 10 (column \mipvierVIBnC ($\epsilon$)). Yet, we are still not able to enumerate the whole frontier for instance E\_20\_9. By seizing the bi-objective character of the model and handing over detected infeasible paths as constraints from one iteration of the $\epsilon$-constraint scheme to the next, we further increase in the algorithms' computational efficiency (\mipvierVIBnCbio($\epsilon$)). Note that different to most works, we add the detected infeasible paths not to a cut pool but explicitly to the set of constraints. All instances with $|P|=20$ can now be solved for \mipvier.
The last two columns of Table~\ref{tab:mip4-comp} show the results obtained by applying the weighting method and conducting a binary search in the objective space. It is again clearly visible, that the approach where cuts are passed on from iteration to another, enhances computation times and thus seizing the bi-objective character of the models is beneficial. Nevertheless, the run times are not comparable to '\mipvierVIBnCbio ($\epsilon$)'. The reason is that the binary search algorithm calls the solver approximately twice as often as the $\epsilon$-constraint.

As noted, instance E\_20\_9 requires significantly more time for computing the Pareto frontier than all the others. The reason is that it is the only instance with $|P|=20$ which has one user with three trips. The total number of trips or average number of trips per person are in line with the other instances. Thus, the maximum number of trips per user has a significant impact. 

Note that we add all found infeasible paths to the set of constraints instead of adding them to a cut pool. As the number of cuts generated is relatively small, and is also decreasing over time, the additional constraints are of a manageable size. However, we have tried both approaches and computational times confirmed the efficiency of our approach.

\begin{table}
  \caption{Average computational times in seconds for each instance of $|P|=20$ solving \mipvier\ using different approaches. The first four sets of results (\mipvier($\epsilon$), \mipvierVI($\epsilon$), \mipvierVIBnC($\epsilon$), \mipvierVIBnCbio($\epsilon$)) are solved using the $\epsilon$-constraint method with either cost (\costfirst) or preferences (\preffirst) set as the objective function. 
  \mipvier($\epsilon$) gives the results by solving \mipvier\ without any additional information, \mipvierVI($\epsilon$) adds valid inequalities to the model, \mipvierVIBnC($\epsilon$) solves the integer model by branch-and-cut, and \mipvierVIBnCbio ($\epsilon$) passes detected infeasible paths as constraints to next iteration. Columns \mipvierVIBnC ($\omega)$ and \mipvierVIBnCbio ($\omega)$ show the results received by the weighting binary search. The latter one again passes former detected infeasible paths as constraints to next subproblems. '\# solved' shows the number of instances solved. TO = time out.}   \label{tab:mip4-comp}%
  \small
  \setlength{\tabcolsep}{2pt}
    \begin{tabular}{lrrrrrrrrrrr}
    \hline\noalign{\smallskip}
          & \multicolumn{2}{c}{\mipvier($\epsilon$)} & \multicolumn{2}{c}{\mipvierVI($\epsilon$)} & \multicolumn{2}{c}{\mipvierVIBnC ($\epsilon$)} & \multicolumn{2}{c}{\mipvierVIBnCbio ($\epsilon$)} & \mipvierVIBnC ($\omega$) & \mipvierVIBnCbio ($\omega$) \\
          & \multicolumn{1}{l}{\costfirst} & \multicolumn{1}{l}{\preffirst} & \multicolumn{1}{l}{\costfirst} & \multicolumn{1}{l}{\preffirst} & \multicolumn{1}{l}{\costfirst} & \multicolumn{1}{l}{\preffirst} & \multicolumn{1}{l}{\costfirst} & \multicolumn{1}{l}{\preffirst} &       &  \\
    \noalign{\smallskip}\hline\noalign{\smallskip}
    E\_20\_0   &  TO    &        26,780  &        12,663  &        7,190 &            779 &       743 &              206  &  \textbf{192}  &    669  &             324  \\
    E\_20\_1   &   TO     &   TO  &    TO  &    TO  &        2,957  &         2,099  &              455  & \textbf{              407} &            1,823 &          621  \\
    E\_20\_2   &  TO  &      20,437 &         13,504  &         9,339  &           737 &        793 &              168  & \textbf{              159} &             608 &               192  \\
    E\_20\_3   &   TO   &    TO  &         29,772  &      18,012 &            856 &          890 &              192  & \textbf{189} &              711 &               294  \\
    E\_20\_4   &   TO   &  TO    &   TO   &        23,490  &     2,447 &         1,798  &              473  &      \textbf{420}  &            1,681  &          560 \\
    E\_20\_5   &   TO   &  TO     &         19,823  &    10,810  &           743  &     781  &              327  & \textbf{              190} &            712  &               322 \\
    E\_20\_6   &  TO     &   TO    &        31,309  &      26,751  &         1,982  &       1,557 &              399  & \textbf{              369} &         1,244 &               590 \\
    E\_20\_7   &  TO    &  TO    &   TO  &    TO  &         3,383 &     2,436  &              628  & \textbf{              496} &     2,555  &               716  \\
    E\_20\_8   &  TO     &   TO   &         32,790 &     18,522  &           892  &        948 &              333  & \textbf{              256} &               729  &               414 \\
    E\_20\_9   &  TO   &   TO   &   TO   &  TO     &    TO    &   TO    &         31,937  & \textbf{        24,629} &  TO   &  TO \\
    \noalign{\smallskip}\hline\noalign{\smallskip}
 \# solved   & 0 & 2 & 6 & 7 & 9 & 9 & 10 & 10 & 9 & 9 \\
   \noalign{\smallskip}\hline
    \end{tabular}%
\end{table}%

The above results show that '\mipvierVIBnCbio ($\epsilon$)' (with direction \preffirst) is, for our problem setting, more efficient than '\mipvierVIBnCbio ($\omega$)'. As discussed, this is mainly due to the increase in the number of MIPs that have to be solved. Table~\ref{tab:mip4-50-comp} compares the run times of the two approaches for $|P|=50$. The table shows similar results as above. The $\epsilon$-constraint method is able to solve more instances and also, if the instance is solved by both approaches, results in shorter computation times. 

\begin{table}
  \caption{Average computational times for each instance for $|P|=50$ solving \mipvierb\ by branch-and-cut embedded in the $\epsilon$-constraint method (\mipvierVIBnCbio ($\epsilon$)) and in the weighting binary search algorithm (\mipvierVIBnCbio($\omega$)). Both approaches add prior detected infeasible paths as constraints to model.}\label{tab:mip4-50-comp}%
    \begin{tabular}{lrrrr}
    \hline\noalign{\smallskip}
       & \multicolumn{1}{l}{\mipvierVIBnCbio ($\epsilon$)} & \multicolumn{1}{l}{\mipvierVIBnCbio ($\omega$)} \\
    \noalign{\smallskip}\hline\noalign{\smallskip}
    E\_50\_0        &            16,712  &            31,100  \\
    E\_50\_1          &  TO   &  TO  \\
    E\_50\_2         &  TO   &  TO  \\
    E\_50\_3          &            13,759  &            22,657  \\
    E\_50\_4       &              5,876  &            13,954  \\
    E\_50\_5      &            41,764  &  TO  \\
    E\_50\_6        &  TO   &  TO  \\
    E\_50\_7       &  TO   &  TO  \\
    E\_50\_8         &              4,746  &              7,907  \\
    E\_50\_9       &  TO   &  TO  \\
    \noalign{\smallskip}\hline
    \end{tabular}%
\end{table}%

\begin{figure}
    \centering
    \subfloat[\mipvierVIBnC ($\epsilon$)]{
    \includegraphics[width=\textwidth]{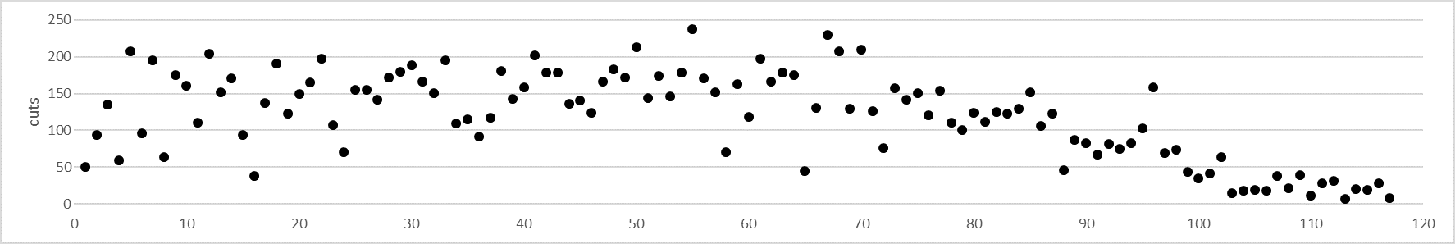}
    }\\
    \subfloat[\mipvierVIBnCbio ($\epsilon$)]{
    \includegraphics[width=\textwidth]{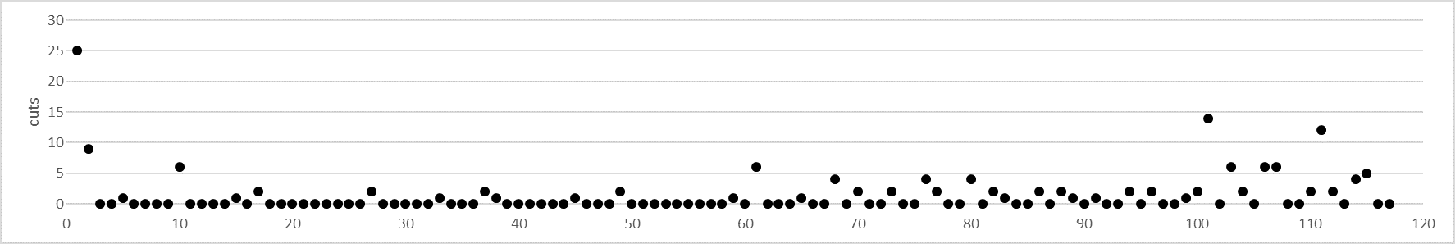}
   }
    \\
    \subfloat[\mipvierVIBnC ($\omega$)]{
    \includegraphics[width=\textwidth]{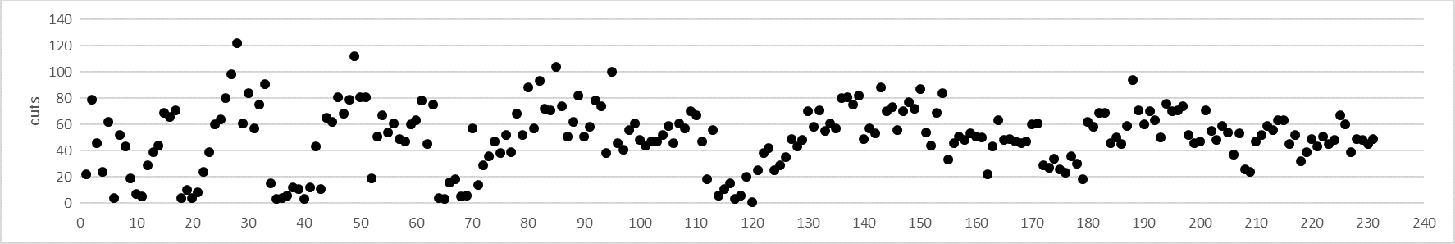}
   }\\
    \subfloat[\mipvierVIBnCbio ($\omega$)]{
    \includegraphics[width=\textwidth]{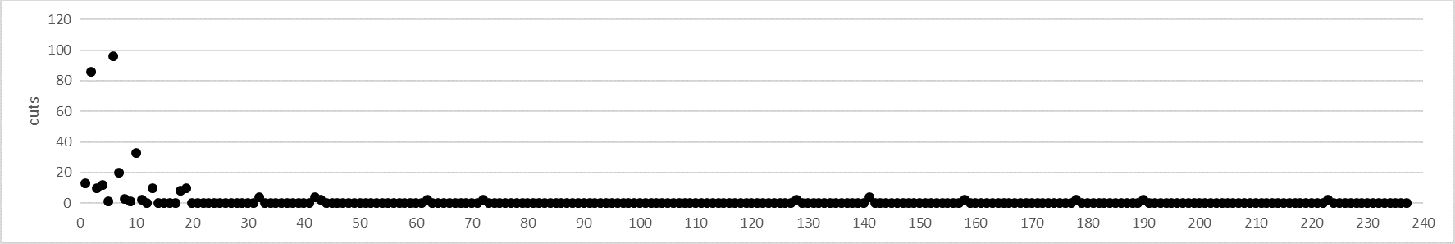}
    }
    \caption{Number of cuts added at each iteration for instance E\_20\_0. \mipvierVIBnC($\epsilon$) solves model \mipvierb\ by branch-and-cut embedded in the $\epsilon$-constraint method, and \mipvierVIBnCbio ($\epsilon$) additionally stores the detected infeasible paths to the set of constraints. \mipvierVIBnC($\omega$) solves model \mipvierb\ by branch-and-cut and the weighting binary search method, and \mipvierVIBnCbio($\omega$) additionally passes infeasible path constraints to subsequent iterations. }
    \label{fig:cutsadded}
\end{figure}

As we have seen, it is beneficial to exploit the bi-objective nature of the underlying optimization problem by using previously generated  cuts in subsequent iterations.
In Figure~\ref{fig:cutsadded}, we show the number of cuts added at each iteration for one chosen instance, namely E\_20\_0. Figures~\ref{fig:cutsadded}(a) and (b) show the results for the $\epsilon$-constraint method, first without adding the cuts as constraints at each iteration and then by using the generated cuts in the respective submodels. Figures~\ref{fig:cutsadded}(c) and (d) give the number of cuts added at each iteration for the weighting method conducting a binary search. As we can see, solving each subproblem individually generates a much higher number of cuts at each iteration, whereas in the other case, where we propagate cuts from iteration to iteration, we drastically reduce the cuts added at each subproblem. This is valid for both methods. Moreover, by comparing Figures~\ref{fig:cutsadded}(b) and (d), we see that the binary search method actually produces fewer cuts in later iterations. 
The reason is that the binary search method detects solutions, where infeasibility needs to be proven. This also results in two times more iterations for this method. 
Nevertheless, we can clearly observe that for either approach, the additional information from prior iterations has a remarkable impact on cut generation iterations.

Table~\ref{tab:cutsadded} gives the number of cuts added per iteration on average for both the $\epsilon$-constraint method as well as the binary search approach for each instance in the set with $|P|=20$. We show the case where each iteration is using only the current information (\mipvierVIBnC ($\epsilon$), \mipvierVIBnC ($\omega$)) and where we use information in the form of cuts added as constraints from prior iterations (\mipvierVIBnCbio($\epsilon$), \mipvierVIBnCbio ($\omega$)). We can clearly see that without additional information we use up to 100 times more cuts. As discussed prior, the binary search method has a lower average number, but more iterations are conducted.

\begin{table}
  \caption{Average number of cuts added at each iteration for instances with $|P|=20$ solving \mipvierb\ by branch-and-cut embedded in the $\epsilon$-constraint method (\mipvierVIBnC($\epsilon$)) or the weighting binary search (\mipvierVIBnC($\omega$)), and by adding detected infeasible paths constraints to the model (\mipvierVIBnCbio ($\epsilon$),\mipvierVIBnCbio ($\omega$))}  \label{tab:cutsadded}%
  \small
  \setlength{\tabcolsep}{2.5pt}
    \begin{tabular}{lrrrrrrrrrr}
    \hline\noalign{\smallskip}
    $|P|=20$ & \multicolumn{1}{c}{E\_20\_0} & \multicolumn{1}{c}{E\_20\_1} & \multicolumn{1}{c}{E\_20\_2} & \multicolumn{1}{c}{E\_20\_3} & \multicolumn{1}{c}{E\_20\_4} & \multicolumn{1}{c}{E\_20\_5} & \multicolumn{1}{c}{E\_20\_6} & \multicolumn{1}{c}{E\_20\_7} & \multicolumn{1}{c}{E\_20\_8} & \multicolumn{1}{c}{E\_20\_9} \\
    \noalign{\smallskip}\hline\noalign{\smallskip}
    \mipvierVIBnC ($\epsilon$) &    120.0  &    205.1  &    180.2  &    223.0  &    227.5  &    179.1  &    208.3  &    189.8  &    179.8  & \multicolumn{1}{r}{   1,694.9 } \\
    \mipvierVIBnCbio($\epsilon$)  &        1.3  &        3.3  &        2.1  &        5.2  &        7.6  &        6.8  &        7.9  &        4.3  &        3.2  & \multicolumn{1}{r}{         70.2 } \\
    \mipvierVIBnC ($\omega$) &      49.8  &    105.0  &      68.6  &      94.1  &    109.7  &      67.3  &      89.6  &      97.7  &      63.0  &   -   \\
    \mipvierVIBnCbio ($\omega$)  &        1.4  &        2.4  &        1.2  &        1.5  &        4.0  &        2.9  &        4.5  &        2.2  &        1.6  &  -  \\
    \noalign{\smallskip}\hline
    \end{tabular}%
\end{table}%

\subsection{Managerial insights}

We briefly discuss managerial implications. We start by looking at the respective Pareto frontiers for a chosen set of instances. Then we continue by studying the different MOT compositions when solving the different variants of the model. Note that the models provide the decision maker with a range of trade-off solutions. Based on this solution pool, the decision maker derives actions and takes the best solution fitting to their requirements.

\subsubsection{Comparison of Pareto frontiers for models \mipeins, \mipzwei, \mipdreiperson, and \mipvier}

\begin{figure}
    \centering
   \subfloat[E\_20\_0]{
    \includegraphics[width=0.75\textwidth]{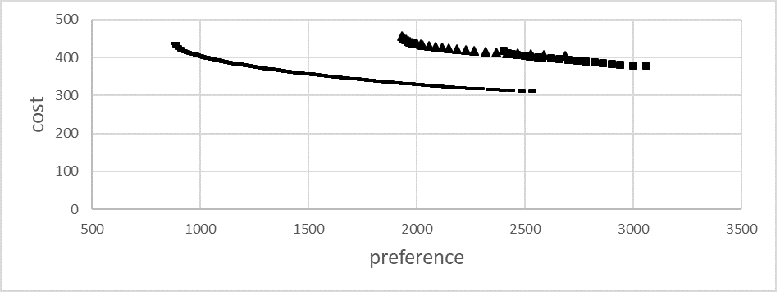}
    } \\
    \subfloat[E\_20\_9]{
    \includegraphics[width=0.75\textwidth]{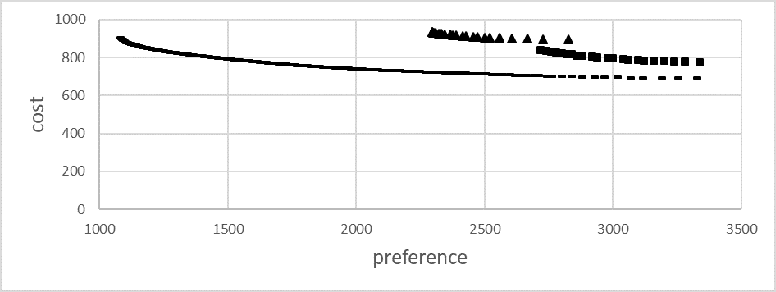}
   }\\
   \subfloat[\mipvierVIBnC ($\epsilon$)]{
    \includegraphics[width=0.3\textwidth]{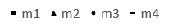}
    }\\
    \captionof{figure}{Pareto frontiers for models \mipeins, \mipzwei, \mipdreiperson, \mipvier\ solving instances E\_20\_0 and E\_20\_9. The y-axis represents cost, preferences are on the x-axis.}
    \label{fig:frontiers}
\end{figure}

Figure~\ref{fig:frontiers} shows the Pareto frontiers for instances E\_20\_0 and E\_20\_9. The x-axis represents preferences, y-axis cost. Note that for both instances only three frontiers are visible. This is because the frontier of \mipeins\ is hidden behind \mipdreiperson. For these small instances, the additional freedom to choose sequences of tasks and trips is not giving any improvement to the model. Frontiers for \mipzwei\ are similar in their shape for both instances, however slightly differ in their relation to the other curves, especially to \mipdreiperson (\mipeins). Introducing time-dependent values for \mipzwei, lower (better) overall preferences but higher cost are obtained, visible as a shift to the left on the x-axis and a shift upwards on the y-axis. The increased cost come from the additional time needed during specific day-times. Note that we usually have a $\beta > 1$, meaning that we rarely decrease the driving time compared to the base scenario (except for public transportation, where we assume shorter cycle times for, e.g., rush-hours). 
For \mipvier, the length of the frontier exceeds all the other curves. It is clearly visible, that with time-dependent preferences and cost as well as flexible sequences, we have a greater set of Pareto optimal solutions. Also, the curve is shifting to the left corner, meaning that we have better overall cost as well as preferences. The average cost and preference values for instances with $u=20$ are: 505 and 2,878 for \mipeins, 548 and 2,272 for \mipzwei, 505 and 2,878 for \mipdreiperson, 476 and 1,591 for \mipvier, respectively. Concluding, we can say that time-dependencies do have a great impact when solving the bi-objective multimodal car-sharing problem. Furthermore, we observe that only dissolving the fixed sequence does not come with high improvements but in combination with time-dependencies a greater amount of solutions as well as lower cost and better user satisfaction is obtained.

\subsubsection{MOT assignment for models \mipeins, \mipzwei, \mipdreiperson, and \mipvier}

\begin{figure}
    \subfloat[E\_20\_0, \mipeins]{
    \includegraphics[width=0.45\textwidth]{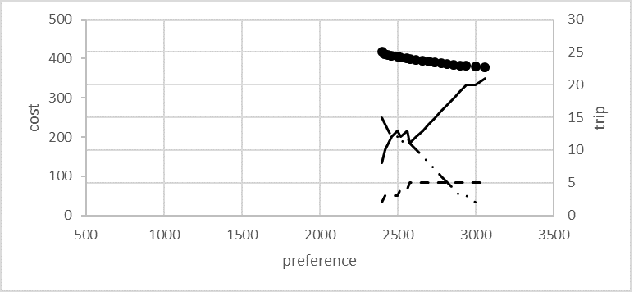}
    }
    \subfloat[E\_20\_9,  \mipeins]{
    \includegraphics[width=0.45\textwidth]{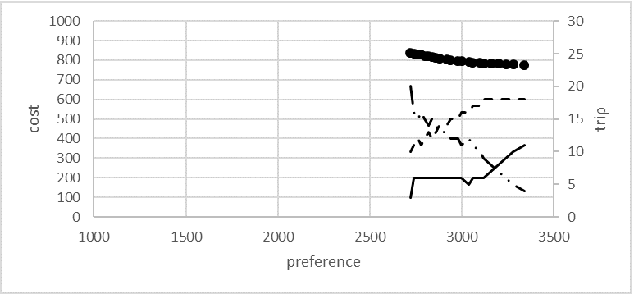}
    }\\
   \subfloat[E\_20\_0, \mipzwei]{
    \includegraphics[width=0.45\textwidth]{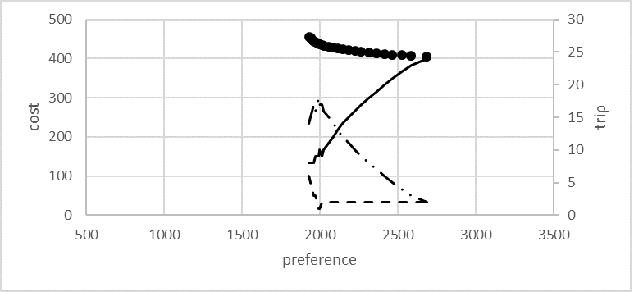}
   }
    \subfloat[E\_20\_9,  \mipzwei]{
    \includegraphics[width=0.45\textwidth]{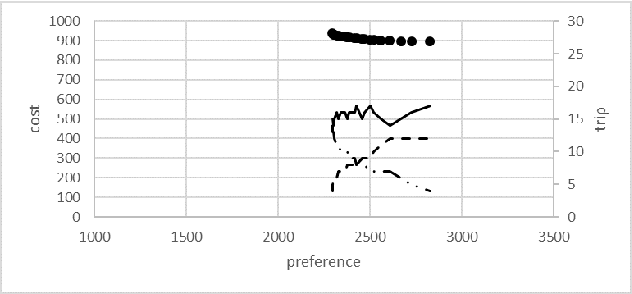}
    }\\
    \subfloat[E\_20\_0, \mipdreiperson]{
    \includegraphics[width=0.45\textwidth]{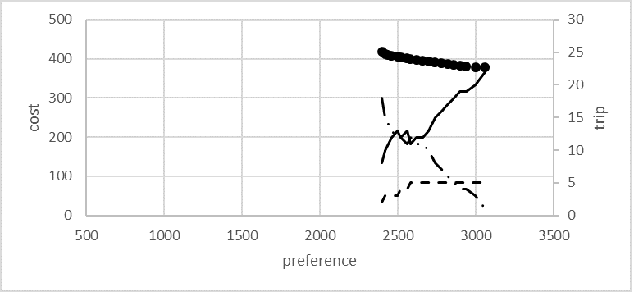}
   }
    \subfloat[E\_20\_9, \mipdreiperson]{
    \includegraphics[width=0.45\textwidth]{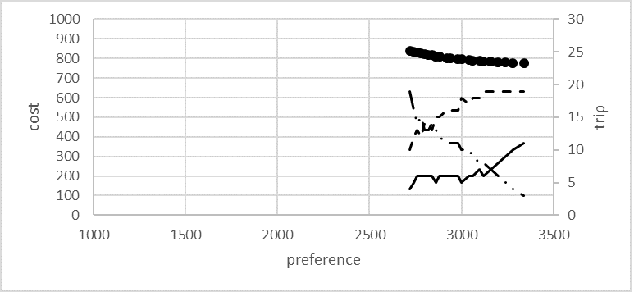}
   }\\
   \subfloat[E\_20\_0,  \mipvier]{
    \includegraphics[width=0.45\textwidth]{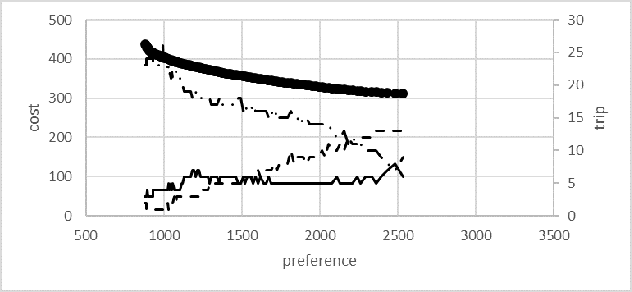}
    }
    \subfloat[E\_20\_9,  \mipvier]{
    \includegraphics[width=0.45\textwidth]{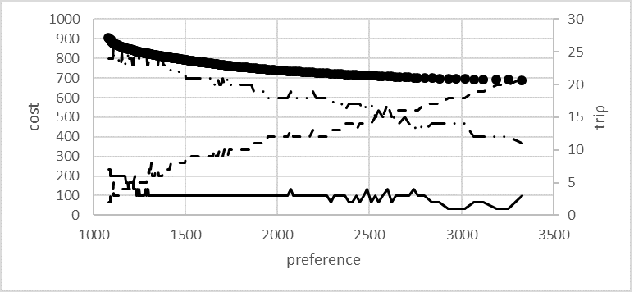}
    }\\
    \centering
    \subfloat[\mipvierVIBnC ($\epsilon$)]{
    \includegraphics[width=0.6\textwidth]{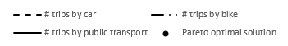}
    }
    \caption{Number of trips assigned for each Pareto optimal solution by the respective mode of transport (car, bike, public transportation) for models \mipeins, \mipzwei, \mipdreiperson, \mipvier\ solving instances E\_20\_0 and E\_20\_9.}
    \label{fig:trips}
\end{figure}

Finally, let us have a closer look at the MOTs assigned. We analyze the number of trips covered by each MOT (car, bike, public transportation), for two instances, namely E\_20\_0 and E\_20\_9, for all four models \mipeins, \mipzwei, \mipdreiperson, \mipvier. In Figure~\ref{fig:trips} we show the respective Pareto frontier for the four models, and include the number of trips taken by each MOT for the respective Pareto optimal solution.
Note that the number of trips that are covered by a car does not have to be equal to the number of cars used in total as a car might take more than one trip during a day. 

Starting with \mipvier, we observe a similar development for both instances for all MOTs. With increasing (worse) preferences, and decreasing cost, we gradually assign more cars and less bikes. The number of trips taken by public transportation is more or less constant. Thus, most cost-efficient, considering time-dependencies, are car trips, best preferences give bike trips. A car is, in our instance set, the fastest mode of transport. As we include time in the cost function, this also makes cars often the cheapest option. Moreover, our study gives relatively good time-dependent preference scores to bikes, as it is, e.g., good in rush-hours to avoid congestion or overcrowded public transportation. This of course, has a great impact on the resulting tendencies in the final results.

For the other models, the picture is slightly different and instance-dependent. Generally, we can say for \mipeins, \mipzwei, and \mipdreiperson\ the number of trips taken by bike is decreasing with lower cost and higher (worse) preferences. The number of trips taken by public transportation are increasing with higher (worse) preferences and lower cost. 

Comparing the extreme points of all Pareto frontiers for all models regarding their composition we can conclude: for \mipeins\ and \mipdreiperson\ we always assign more cars and public transportation to the cost optimal solution (except for one instance for \mipdreiperson), the number of trips taken by public transportation and cars decreases with higher cost but better preferences. Bikes are preferred by the preference optimal solutions, and increase with less cost. 
Also for \mipzwei\ we can observe that the number of bikes assigned is decreasing with increasing cost and lower (better) preferences.
The opposite holds for public transportation. 
We can figure an unchanged level of trips being assigned to public transportation for \mipvier. For \mipvier\, lower cost and higher (worse) preferences lead to more cars assigned and, conversely, more bikes are assigned with an increase in cost, and decrease in preferences. 

Table~\ref{tab:trips-MOTs} provides a better overview of the MOTs assigned to trips for each instance set and model. The numbers are given as averages over all instances within an instance set. Rows 'av' provide the average of the average number of trips by the respective MOT (car, bike, public). 'min' gives the average of the minimum number of trips conducted by the respective MOT, and 'max' gives the average maximum number. The results are organised by model (\mipeins, \mipzwei, \mipdreiperson, \mipvier), MOTs, and number of users $|P|=20,50,100,150,200,250,300$.

Generally, we observe that bikes are very often assigned and used for the highest number of trips on average. \mipdreiperson\ assigns about the same amount of cars and public transportation. \mipzwei\ always shows the highest number of trips taken by public transportation. Thus, by having the choice between MOTs for a trip with a fixed sequence, public transportation is preferred.  \mipvier\ has a very high number of trips taken by bikes. 

Note that the composition of the mobility offers varies a lot among the models. Furthermore, the difference between the minimums and maximums of the assigned MOTs is usually very high, which means that the solutions are changing considerably over the course of the Pareto frontier. This means that, from a decision makers perspective, considering the proposed trade-offs and variants of the problem has a big impact on the MOTs used in a mobility system. Assigning different MOTs influences the user-centred objective to a great extent. With this results we can confirm the relevance of this study and conclude that it is highly beneficial to consider not only cost but also user-preferences when operating a shared mobility system. 

\begin{table}
  \caption{Average values of average number of MOT assigned to trips (av), minimum (min) or maximum (max) for models \mipeins, \mipzwei, \mipdreiperson, \mipvier\ for an increasing number of users $|P|=20,50,100,150,200,250,300$. Considered modes of transport are car, bike and public transportation.}  \label{tab:trips-MOTs}%
  \setlength{\tabcolsep}{3.5pt}
  \small
    \begin{tabular}{lc|rrr|rrr|ccc|ccc}
    \hline\noalign{\smallskip}
          &       & \multicolumn{3}{c|}{\mipeins} & \multicolumn{3}{c|}{\mipzwei} & \multicolumn{3}{c|}{\mipdreiperson} & \multicolumn{3}{c}{\mipvier} \\
          &   $|P|$    & \multicolumn{1}{l}{car} & \multicolumn{1}{l}{bike} & \multicolumn{1}{l|}{public} & \multicolumn{1}{l}{car} & \multicolumn{1}{l}{bike} & \multicolumn{1}{l|}{public} & \multicolumn{1}{l}{car} & \multicolumn{1}{l}{bike} & \multicolumn{1}{l|}{public} & \multicolumn{1}{l}{car} & \multicolumn{1}{l}{bike} & \multicolumn{1}{l}{public} \\
    \noalign{\smallskip}\hline\noalign{\smallskip}
    av    & \multirow{3}[2]{*}{20} &        7.7  &      11.7  &      11.4  &      2.2  &      11.6  &      17.1  & \multicolumn{1}{r}{       8.1 } & \multicolumn{1}{r}{     10.5 } & \multicolumn{1}{r|}{    12.1 } & \multicolumn{1}{r}{          6.3 } & \multicolumn{1}{r}{   20.0 } & \multicolumn{1}{r}{      4.5 } \\
    min   &       &        4.5  &        4.7  &        6.8  &      0.8  &        5.0  &      11.5  & \multicolumn{1}{r}{       4.7 } & \multicolumn{1}{r}{       3.9 } & \multicolumn{1}{r|}{      7.5 } & \multicolumn{1}{r}{          1.8 } & \multicolumn{1}{r}{   12.1 } & \multicolumn{1}{r}{      2.5 } \\
    max   &       &      10.6  &      19.3  &      15.9  &      4.5  &      16.8  &      23.6  & \multicolumn{1}{r}{     10.9 } & \multicolumn{1}{r}{     18.3 } & \multicolumn{1}{r|}{    16.4 } & \multicolumn{1}{r}{       14.6 } & \multicolumn{1}{r}{   26.3 } & \multicolumn{1}{r}{      6.8 } \\
    \noalign{\smallskip}\hline\noalign{\smallskip}
    av    & \multirow{3}[2]{*}{50} &      23.4  &      29.9  &      23.0  &      3.1  &      27.9  &      45.3  & \multicolumn{1}{r}{     23.3 } & \multicolumn{1}{r}{     27.6 } & \multicolumn{1}{r|}{    25.3 } &    21.6 
       &   45.0    &    10.0    \\
    min   &       &      11.0  &      11.5  &      14.4  &      1.8  &      11.2  &      29.1  & \multicolumn{1}{r}{     11.8 } & \multicolumn{1}{r}{       9.1 } & \multicolumn{1}{r|}{    16.4 } &           5.3 	   &   29.8    &      5.1   \\
    max   &       &      33.9  &      50.6  &      32.3  &      6.5  &      43.6  &      61.1  & \multicolumn{1}{r}{     34.9 } & \multicolumn{1}{r}{     46.2 } & \multicolumn{1}{r|}{    34.9 } &     41.0 
      &   62.0  &   14.5     \\
    \noalign{\smallskip}\hline\noalign{\smallskip}
    av    & \multirow{3}[2]{*}{100} &      47.1  &      54.2  &      45.3  &      5.9  &      54.2  &      86.5  & \multicolumn{1}{r}{     47.3 } & \multicolumn{1}{r}{     54.0 } & \multicolumn{1}{r|}{    45.4 } &  -    &  -    &  -  \\
    min   &       &      22.5  &      20.8  &      30.1  &      2.9  &      19.9  &      54.1  & \multicolumn{1}{r}{     23.8 } & \multicolumn{1}{r}{     19.9 } & \multicolumn{1}{r|}{    31.0 } &  -    &  -    &  -  \\
    max   &       &      71.7  &      93.9  &      57.0  &    13.2  &      82.8  &    120.0  & \multicolumn{1}{r}{     72.0 } & \multicolumn{1}{r}{     87.9 } & \multicolumn{1}{r|}{    61.4 } &  -    &  -    &  -  \\
    \noalign{\smallskip}\hline\noalign{\smallskip}
    av    & \multirow{3}[2]{*}{150} &      72.7  &      82.7  &      62.2  &    10.8  &      81.4  &    125.4  &      66.4  &      86.9  &     64.8  &  -    &  -    &  -  \\
    min   &       &      37.5  &      34.3  &      41.8  &      6.4  &      32.7  &      82.3  &      33.5  &      35.5  &     45.6  &  -    &  -    &  -  \\
    max   &       &    106.1  &    138.2  &      79.4  &    21.0  &    121.0  &    171.4  &    102.4  &    136.5  &     86.4  &  -    &  -    &  -  \\
    \noalign{\smallskip}\hline\noalign{\smallskip}
    av    & \multirow{3}[2]{*}{200} &      94.6  &    108.6  &      83.9  &    15.4  &    107.4  &    164.3  &  -    &  -    &  -    &  -    &  -    &  -  \\
    min   &       &      48.5  &      39.5  &      55.1  &    10.3  &      40.1  &    105.4  &  -    &  -    &  -    &  -    &  -    &  -  \\
    max   &       &    142.1  &    183.3  &    107.0  &    25.3  &    163.1  &    227.3  &  -    &  -    &  -    &  -    &  -    &  -  \\
    \noalign{\smallskip}\hline\noalign{\smallskip}
    av    & \multirow{3}[2]{*}{250} &    120.0  &    137.0  &    101.2  &    20.9  &    137.1  &    200.3  &  -    &  -    &  -    &  -    &  -    &  -  \\
    min   &       &      62.2  &      53.0  &      65.3  &    14.3  &      51.5  &    129.5  &  -    &  -    &  -    &  -    &  -    &  -  \\
    max   &       &    177.7  &    230.4  &    130.4  &    37.7  &    201.8  &    279.6  &  -    &  -    &  -    &  -    &  -    &  -  \\
    \noalign{\smallskip}\hline\noalign{\smallskip}
    av    & \multirow{3}[2]{*}{300} &    131.2  &    163.8  &    132.5  &    19.6  &    163.2  &    244.7  &  -    &  -    &  -    &  -    &  -    &  -  \\
    min   &       &      70.3  &      62.7  &      88.4  &    13.8  &      57.8  &    158.5  &  -    &  -    &  -    &  -    &  -    &  -  \\
    max   &       &    194.7  &    268.7  &    171.9  &    36.6  &    239.7  &    347.3  &  -    &  -    &  -    &  -    &  -    &  -  \\
    \noalign{\smallskip}\hline
    \end{tabular}%
\end{table}%

\section{Conclusion and future work} \label{sec:concl}

Inspired by the change in mobility patterns we study the bi-objective multimodal car-sharing problem where we assign modes of transport to trips as well as cars and user routes. As objectives we consider costs and user-centred preferences. Both objectives are, depending on the variant of the model, studied with time dependencies. We model different cost/times as well as preferences during a day, as people might want to avoid driving through rush-hour by car. We introduce four different variants of the model where we gradually dissolve a fixed sequence of tasks and trips as well as introduce the effect of the time-dependent values. The increase in flexibility in the model comes with an increase in the complexity as well as a an increase in the number of Pareto optimal solutions. Therefore, we reformulate the last variant, without fixed sequences and time-dependencies, to a purely integer model and propose a branch-and-cut algorithm. We show that our branch-and-cut algorithm can enumerate the Pareto frontier for prior non-tractable instances within seconds. We embed the algorithm into two bi-objective frameworks, namely the $\epsilon$-constraint method and a weighting binary search method. We show that adding previously detected infeasible path constraints to subsequent iterations reduces computational times considerably. In our computational study we observe that only dissolving the fixed sequence does not come with high improvements. However, in combination with time-dependencies a greater amount of solutions as well as lower cost and better user satisfaction is obtained. Moreover, we observe that the solutions change significantly along the Pareto frontier. This confirms the relevance of this study. We conclude that it is highly beneficial to consider not only cost but also user-preferences when operating a shared mobility system. 

Even though we are able to show a significant enhancement in computational efficiency for a set of instances, our approach has limitations. Enumerating the whole Pareto frontier for instances with users having more than two trips throughout a day, seems challenging. Thus, future work should tackle this issue by focusing on the development of a separation algorithm adjusted to these specific characteristics. Moreover, specific matheuristics where the relative MIP-gap is increased or the $\epsilon$ value adapted, may lead to promising further improvement in run times. Furthermore, the development of metaheuristics should enable an increase in computational efficiency for the proposed problem. Finally, as this work only optimizes average scores of preferences, a min-max approach is planned for future work in order to improve the integration of preferences on a user level.

\section*{Acknowledgements}
This work has been partially funded by the Climate and Energy Funds (KliEn) under grant number 853767. This research was funded in whole, or in part, by the Austrian Science Fund (FWF) [P 31366]. For the purpose of open access, the author has applied a CC BY public copyright licence to any Author Accepted Manuscript version arising from this submission.

%
\section*{Conflict of interest}

The authors declare that they have no conflict of interest.

\bibliographystyle{spbasic}      
\bibliography{BiO-MMCP_arxiv}

\newpage

\pagenumbering{Roman}
\setcounter{page}{1}

\section{Appendix}

\begin{algorithm}[H]
\SetAlgoLined
\small
\ForAll{$p \in P$}{
construct empty route vector $\rho$\;
 $v = \gamma_p$\;
 \While{$v \neq \phi_p$}{
    \ForAll{{$l$} $\in$ v.outgoing}{
        if({$l$}.person = $p$ \& $\overline{x}_l$ = 1) 
        store leg $l$ in vector $\rho$\;
        $v$ = {$l$}.endNode\;
        break\;
    }
 }
 $\Delta = \infty$; $\tau = 0$; $\mathcal{W}$ = 0; F = 0, {$l$} = 1\;
 \ForAll{$l \leq |\rho_{l}|$}{
 if ({$l$}.startNode = trip start node $a$) $\Delta = \infty$; $F = 0$; $\mathcal{W}$ = 0\;
 $\tau$ = $\tau$ + $s_{v}$ + $t_{l-1}$\;
 $\mathcal{W}$ ${\mathrel{+}=}$ min\{max\{0,$e_{l}$-$\tau$\},h\}\;
 $\Delta$ = min\{$\Delta$, max\{0,$e_{l}$-($\tau$+$\mathcal{W}$)\}\}\;
F = $\mathcal{W}$ + $\Delta$\;
 \eIf{$o_{l} \leq \tau \leq e_{l}$}{
    {$l$}++\;
    }{
    $\tau'$ = $\tau$\;
    $\tau$ ${\mathrel{+}=}$ min\{max\{0,$o_{l}-\tau$\}, F\}\;
    \eIf{$o_{l} \leq \tau \leq e_{l}$}{
        $\Delta$ ${\mathrel{+}=}$ min\{$\mathcal{W}$ - ($\tau-\tau'$),0\}\;
         $\mathcal{W}$ = max\{$\mathcal{W}$ - ($\tau-\tau'$),0\}\;
         {$l$}++;
        }{
        add Cut (see Section~\ref{sec:cutsadding});
        }
    }
 }
}
\caption{Separation of user routes.}
\label{algo:user}
\end{algorithm}

\begin{algorithm}[H]
\SetAlgoLined
\small
\ForAll{$m \in K: m = car$}{
trips = 0\;
 \ForAll{$v \in D$}{
    \ForAll{{$l$} $\in$ v.outgoing}{
    if ({$l$}.mot $\neq$ m) continue\;
    construct empty vector $\rho$\;
        if($\overline{x}_l$ = 1) 
        store leg $l$ in vector $\rho$\;
        $v$ = {$l$}.endNode\;
        \While{$v \neq carEndNode $}{
         \ForAll{{$l$} $\in$ v.outgoing}{
         if ({$l$}.mot != m) continue\;
         if($\overline{x}_l$ = 1)
         store leg $l$ in vector $\rho$\;
         $v$ = {$l$}.endNode\;
         if ($v$ = startNode) trips++\;
         break\;
         }
        }
 if (trips $<$ 2) continue\;
 $\Delta = \infty$; $\tau = 0$; $\mathcal{W}$ = 0; F = 0, {$l$} = 1\;
 \ForAll{$l \leq |\rho_{l}|$}{
 if ({$l$}.startNode = trip start node $a$) $\Delta = \infty$; $F = 0$; $\mathcal{W}$ = 0\;
 $\tau$ = $\tau$ + $s_{l.startNode}$ + $t_{l-1}$\;
 $\mathcal{W}$ ${\mathrel{+}=}$ min\{max\{0,$e_{l}$-$\tau$\},h\}\;
 $\Delta$ = min\{$\Delta$, max\{0,$e_{l}$-($\tau$+$\mathcal{W}$)\}\}\;
F = $\mathcal{W}$ + $\Delta$\;
 \eIf{$o_{l} \leq \tau \leq e_{l}$}{
    {$l$}++\;
    }{
    $\tau'$ = $\tau$\;
    $\tau$ ${\mathrel{+}=}$ min\{max\{0,$o_{l}-\tau$\}, F\}\;
    \eIf{$o_{l} \leq \tau \leq e_{l}$}{
        $\Delta$ ${\mathrel{+}=}$ min\{$\mathcal{W}$ - ($\tau-\tau'$),0\}\;
         $\mathcal{W}$ = max\{$\mathcal{W}$ - ($\tau-\tau'$),0\}\;
         {$l$}++;
        }{
        add Cut (see Section~\ref{sec:cutsadding});
        }
    }
 }
}
    }
 }
\caption{Separation of car routes.}
\label{algo:vehicle}
\end{algorithm}

\begin{table}
  \caption{Preference scores assigned based on the binary assignment from the instance generation. }   \label{tab:preferencescore-base}%
    \begin{tabular}{rrrrrrrrr}
    \hline\noalign{\smallskip}
    \multicolumn{5}{c}{binary assignment of MOTs} &       & \multicolumn{3}{c}{preference scores} \\
    \multicolumn{1}{l}{walk} & \multicolumn{1}{l}{bike} & \multicolumn{1}{l}{car} & \multicolumn{1}{l}{e-car} & \multicolumn{1}{l}{public} &       & \multicolumn{1}{l}{walk/public} & \multicolumn{1}{l}{bike} & \multicolumn{1}{l}{car} \\
    \noalign{\smallskip}\hline\noalign{\smallskip}
    1     & 0     & 0     & 0     & 0     &       & 4     & 6     & 7 \\
    0     & 1     & 0     & 0     & 0     &       & 6     & 4     & 7 \\
    0     & 0     & 0     & 0     & 1     &       & 4     & 6     & 7 \\
    0     & 0     & 1     & 0     & 0     &       & 6     & 7     & 4 \\
    0     & 0     & 0     & 1     & 0     &       & 6     & 7     & 5 \\
    1     & 1     & 0     & 0     & 0     &       & 4     & 4     & 7 \\
    0     & 1     & 0     & 0     & 1     &       & 4     & 4     & 7 \\
    0     & 0     & 1     & 0     & 1     &       & 4     & 5     & 4 \\
    0     & 0     & 1     & 1     & 0     &       & 7     & 7     & 4 \\
    1     & 0     & 0     & 0     & 1     &       & 4     & 6     & 7 \\
    0     & 1     & 1     & 0     & 0     &       & 6     & 4     & 4 \\
    0     & 0     & 0     & 1     & 1     &       & 4     & 7     & 5 \\
    1     & 0     & 1     & 0     & 0     &       & 4     & 5     & 4 \\
    0     & 1     & 0     & 1     & 0     &       & 6     & 5     & 6 \\
    1     & 1     & 0     & 0     & 1     &       & 4     & 4     & 7 \\
    1     & 0     & 0     & 1     & 0     &       & 4     & 7     & 5 \\
    0     & 1     & 1     & 0     & 1     &       & 4     & 4     & 4 \\
    0     & 0     & 1     & 1     & 1     &       & 7     & 7     & 4 \\
    1     & 1     & 1     & 0     & 1     &       & 4     & 4     & 4 \\
    0     & 1     & 1     & 1     & 1     &       & 7     & 4     & 4 \\
    1     & 0     & 1     & 1     & 1     &       & 4     & 7     & 4 \\
    1     & 1     & 1     & 1     & 0     &       & 4     & 4     & 4 \\
    1     & 1     & 0     & 1     & 1     &       & 4     & 4     & 7 \\
    1     & 1     & 1     & 1     & 1     &       & 4     & 4     & 4 \\
    0     & 0     & 0     & 0     & 0     &       & 4     & 5     & 5 \\
    \noalign{\smallskip}\hline
    \end{tabular}%
\end{table}%

\begin{table}
  \caption{On the left: Adaption of the user preferences for the time-dependent values for each time period $t$. The base values are taken from Table~\ref{tab:preferencescore-base} and accordingly deducted/added.
  On the right: $\beta$-values to multiply the respective cost and time value of the respective MOT for the respective time periods $t$.}  \label{tab:preferencescore-period}%
    \begin{tabular}{crrr}
    \hline\noalign{\smallskip}
     $t$     & \multicolumn{1}{l}{walk/public} & \multicolumn{1}{l}{bike} & \multicolumn{1}{l}{car} \\
    \noalign{\smallskip}\hline\noalign{\smallskip}
    0     & -3    & -2    & +1 \\
    1     & +2    & -2    & +3 \\
    2     & -2    & -1    & -3 \\
    3     & +\- 0    & +\- 0    & +\- 0 \\
    4     & -2    & -3    & +1 \\
    5     & +2    & -2    & +3 \\
    6     & -1    & +2    & -2 \\
   \noalign{\smallskip}\hline
    \end{tabular}%
      \begin{tabular}{cc}
      &  
    \end{tabular}%
      \begin{tabular}{rrr}
    \hline\noalign{\smallskip}
     \multicolumn{1}{c}{car} & \multicolumn{1}{l}{walk/public} & \multicolumn{1}{l}{bike}  \\
   \noalign{\smallskip}\hline\noalign{\smallskip}
     1.2   & 1.1 & 1 \\
     1.4   & 0.8  & 1.1\\
     1.3   & 0.9 & 1 \\
     1      &   1  & 1\\
     1 .3  & 1  & 1\\
    1.4   & 0.9 & 1.1 \\
     1.1   & 1.3  & 1\\
    \noalign{\smallskip}\hline
    \end{tabular}%
  \label{tab:costtime-factor}%
\end{table}%

\end{document}